\documentclass{article}

\usepackage[preprint,nonatbib]{neurips_2024}

\usepackage[utf8]{inputenc} %
\usepackage[T1]{fontenc}    %
\usepackage{hyperref}       %
\usepackage{url}            %
\usepackage{booktabs}       %
\usepackage{amsfonts}       %
\usepackage{nicefrac}       %
\usepackage{microtype}      %
\usepackage{xcolor}         %
\usepackage{bm}
\usepackage{multirow}

\usepackage{graphicx}
\usepackage{tikz}
\usepackage{caption}
\usepackage{subcaption}
\usetikzlibrary{positioning}
\definecolor{GHOST}{HTML}{FFFFFF}

\newcommand{\pmh}[1]{\textcolor{violet}{[\textit{#1}]}}
\newcommand{\ta}[1]{\textcolor{blue}{[\textit{#1}]}}
\newcommand{\di}[1]{\textcolor{orange}{[\textit{#1}]}}

\renewcommand{\pmh}[1]{}
\renewcommand{\ta}[1]{}
\renewcommand{\di}[1]{}

\newcommand{\neuripsonly}[1]{}

\renewcommand{\paragraph}[1]{\par\noindent\textbf{#1}~}

\title{Sampling 3D Gaussian Scenes in Seconds with\\Latent Diffusion Models}

\author{%
  Paul Henderson \textsuperscript{\textnormal{1}} ~~~
  Melonie de Almeida \textsuperscript{\textnormal{1}} ~~~
  Daniela Ivanova \textsuperscript{\textnormal{1}} ~~~
  Titas Anciukevičius \textsuperscript{\textnormal{2}} \\
~\textsuperscript{1} University of Glasgow ~
\textsuperscript{2} University of Edinburgh ~%
}

\begin{document}

\maketitle

\begin{abstract}
We present a latent diffusion model over 3D scenes, that can be trained using only 2D image data.
To achieve this, we first design an autoencoder that maps multi-view images to 3D Gaussian splats, and simultaneously builds a compressed latent representation of these splats.
Then, we train a multi-view diffusion model over the latent space to learn an efficient generative model.
This pipeline does not require object masks nor depths, and is suitable for complex scenes with arbitrary camera positions.
We conduct careful experiments on two large-scale  datasets of complex real-world scenes -- MVImgNet and RealEstate10K.
We show that our approach enables generating 3D scenes in as little as 0.2 seconds, either from scratch, from a single input view, or from sparse input views.
It produces diverse and high-quality results while running an order of magnitude faster than non-latent diffusion models and earlier NeRF-based generative models.
\end{abstract}

\section{Introduction}

Learning generative models that capture the distribution of the 3D world around us is a compelling yet challenging problem.
As well as the grander aim of building intelligent agents that can understand their environment, such models are also useful for many practical tasks.
In games and visual effects, they enable effortless creation of 3D assets, which currently is notoriously difficult, slow and expensive. 
In computer vision, they enable 3D reconstruction of realistic scenes from a single image, with the generative model synthesising plausible 3D details even for regions not visible in the image---unlike classical 3D reconstruction methods \cite{hartley2003multiple,mildenhall2020nerf}.

Large-scale datasets of images, text and video \cite{schuhmann2022laion, gao2020pile, blattmann2023stable} have enabled learning impressive generative models for those modalities \cite{Rombach_2022_CVPR, radford2018improving, ho2022imagen}.
However, there are no extant large-scale datasets of photorealistic 3D scenes.
Existing 3D datasets are either large but consist primarily of isolated objects (not full scenes), often with unrealistic textures \cite{wu2023omniobject3d,deitke2024objaverse, chang2015shapenet}; or they are photorealistic environments (captured with 3D scanners) but too small for learning a generative model over \cite{dai2017scannet,armeni2017joint}.
In contrast, large-scale in-the-wild datasets of \textit{multi-view images} are now readily available \cite{yu2023mvimgnet,zhou2018stereomag,reizenstein21co3d}.

It is therefore desirable to learn 3D generative models directly from datasets of multi-view images, rather than from 3D data.
One na\"ive strategy is to apply standard 3D reconstruction techniques to every scene in such a dataset, then train a 3D generative model directly on the resulting reconstructions \cite{mueller2022diffrf,zhang2024gaussiancube}.
However, this is computationally expensive.
It also leads to a challenging learning task for the generative model, since by reconstructing scenes independently we do not obtain a smooth, shared space of representations (e.g.~similar scenes may have very different weights when represented as a NeRF~\cite{mildenhall2020nerf}).
This makes it difficult to learn a prior that generalises, rather than simply memorising individual scenes.
These limitations have inspired a line of works that learn 3D generative models directly from images \cite{anciukevicius2022renderdiffusion,szymanowicz2023viewset_diffusion,schwarz2020graf,henderson20cvpr}.
Unfortunately, recent methods are very slow to sample, since they require expensive volumetric rendering operations after every step of a diffusion process \cite{szymanowicz2023viewset_diffusion,anciukevicius2024denoising,tewari2023forwarddiffusion}.

In this work, we design an efficient generative model for 3D scenes that is trained using only posed multi-view images.
Our key idea is to learn an autoencoder on multi-view images, that simultaneously builds a structured 3D representation while also compressing this into a lower dimensionality latent space.
We can then train a denoiser on the resulting latent representations, and perform the expensive iterative denoising process in this much lower-dimensional space.
Unlike previous 3D-aware diffusion models \cite{szymanowicz2023viewset_diffusion,tewari2023forwarddiffusion}, the rendering operation is \textit{not} inside the sampling loop.

We adopt Gaussian Splats \cite{kerbl20233d} as our 3D representation.
Recent work on 3D reconstruction has shown splats achieve a favorable trade-off between reconstruction quality and training/rendering speed.
The original work of Kerbl \textit{et al.}~\cite{kerbl20233d} considered only reconstruction from densely-captured images, but several more recent works aim to predict splats from one or few images \cite{chen2024mvsplat,charatan2023pixelsplat,szymanowicz2024splatter_image,gslrm2024, xu2024grm}.
However, unlike ours, these methods are not generative -- they do not capture a \textit{distribution} over 3D scenes.
They therefore cannot predict the full space of scenes consistent with the input images; nor can they perform other tasks such as unconditional or class-conditional generation.
Score distillation \cite{poole2022dreamfusion,wang2022score} provides a way to leverage 2D generative models for 3D content synthesis, but such methods neither learn nor sample a true 3D prior, and so are still prone to undesirable artifacts \cite{tang2023dreamgaussian, yi2023gaussiandreamer,chung2023luciddreamer}.

Our proposed model is very fast, since
it benefits from the efficient rendering and optimisation of Gaussian Splats, and also from the speed-up provided by diffusion over a compressed latent space.
This enables sampling a full batch of eight 3D scenes in just 1.6s---more than $20\times$ faster than the fastest existing 3D-aware diffusion model \cite{szymanowicz2023viewset_diffusion}.
In addition, our model has the following desirable features: 
(i) it can represent arbitrarily large scenes, by placing 3D content anywhere in the view frustum of the cameras (in particular, it is not limited to pre-segmented or object-centric scenes);
(ii) it supports several tasks---unconditional generation, single-image 3D reconstruction, sparse-view 3D reconstruction, depending on the conditioning signal provide during inference;
(iii) it does not rely on any depth or segmentation estimates or annotations for training---it can be trained from scratch using only posed multi-view images.

To summarise, our core contribution is \textbf{the first generative model that learns and samples a distribution over real-world scenes represented as Gaussian Splats}.
Our technical contributions that enable this are:
    (i) we design 
    a new 3D-aware autoencoder architecture that learns to represent multi-view images via a compressed latent space, that can be decoded to Gaussian Splats;
    (ii) we demonstrate how diverse and realistic 3D scenes can be sampled efficiently with a diffusion model on the latent representation, either unconditionally or conditioned on an input image;
    (iii) we show that for a given compute budget, our latent approach gives significantly better results for both unconditional generation and generative reconstruction.

\section{Related Work}

\paragraph{Reconstruction from dense views.}
Numerous scene representations and corresponding inference methods have been proposed, including surface representations (e.g.~meshes, distance fields \cite{park2019deepsdf}), point clouds \cite{schoenberger2016sfm, seitz2006comparison, snavely2006photo}, light fields \cite{curless1996volumetric, gortler1996lumigraph} and volumetric representations (e.g.~radiance fields~\cite{xie22neuralfields} and voxels \cite{seitz1999photorealistic, sitzmann2019deepvoxels}).
Current state-of-the-art methods use neural radiance fields (NeRFs) \cite{mildenhall2020nerf, barron2021mip}, which implicitly parameterize a radiance field with a neural network, making them easily optimizable with gradient descent by minimizing an image reconstruction loss.
However, NeRFs require expensive volumetric rendering involving numerous MLP queries, and despite recent efforts to reduce their size \cite{peng2020convolutional, fridovich2023k, anciukevicius2024denoising, muller2022instant, li2023neuralangelo, chen2022tensorf, xu2022point, fridovich2023k}, both training and rendering remain slow.
Recently, Gaussian Splatting \cite{kerbl20233d} was introduced as an alternative that allows real-time rendering and fast training, with quality approaching that of state-of-the-art NeRFs. 
Our work also uses this efficient representation, but instead of fitting individual scenes, we build a generative model that learns to sample them from a distribution (e.g.~conditioned on a class label or sparse set of images).

\paragraph{Reconstruction from sparse views.}
While the above methods can reconstruct 3D from dense (e.g.~>50) sets of images, in practice, parts of a scene cannot be observed from multiple images and need to inferred.
To solve this, a line of work trains models to reconstruct 3D scenes from fewer (e.g.~<10) views. Most approaches \cite{yu2021pixelnerf, wang2021ibrnet, mvsnerf, liu2022neural, henzler2021unsupervised, wiles2020synsin, liu2022neural, wu2023multiview, rockwell2021pixelsynth} unproject 2D image features into 3D space, fuse them and apply NeRF rendering.
Several recent and concurrent methods \cite{charatan2023pixelsplat, chen2024mvsplat, zou2023triplane, gslrm2024, xu2024grm, zheng2024gpsgaussian, wewer24latentsplat, shen2024gamba, szymanowicz2024splatter_image} tackle the sparse-view reconstruction task using splats as the 3D representation; others aim to predict novel views directly, without explicit 3D \cite{kulhanek2022viewformer, rombach2021geometryfree, du2023cross}.
However, these methods are not probabilistic---they do not represent uncertainty about unobserved parts of the scene (e.g.~the back of an object). %
As a result, instead of sampling one of many plausible 3D representations, these methods output a single average solution.
For example, \textit{MVSplat}~\cite{chen2024mvsplat} and \textit{pixelSplat}~\cite{charatan2023pixelsplat} train a network to map context images to 3D splats, but lack the ability to generate 3D scenes unconditionally or to sample diverse completions for occluded regions.
\textit{latentSplat} \cite{wewer24latentsplat} does construct a posterior distribution over splats then sample this---however it imposes a mean-field (independent) posterior on the splat parameters themselves, meaning it cannot capture complex posterior dependencies in the scene geometry, and thus cannot sample coherent shapes for unobserved areas of scenes.
Note that almost all the above methods are designed for two or more input views. Only SplatterImage~\cite{szymanowicz2024splatter_image} can predict 3D splats from a single image, and even this method is limited to masked, object-centric scenes.
While some approaches incorporate information from generative models ad-hoc to increase plausibility of uncertain regions \cite{shriram2024realmdreamer, roessle2023ganerf, zou2023sparse3d, melaskyriazi2023realfusion, chung2023luciddreamer, regnerf2022}, they do not learn to sample the true posterior distribution on scenes, which our method aims to learn.

\begin{figure}
    \centering
    \includegraphics[width=\linewidth]{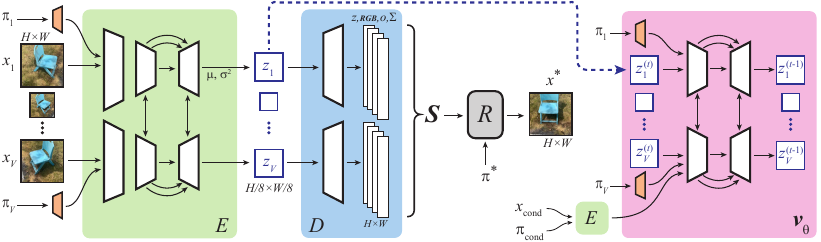}
    \caption{Overview of our latent diffusion model for 3D scene synthesis.
    \textbf{Left:} We train an autoencoder, that encodes (green box; $E$) multi-view images $\{x_v\}_1^V$ to a compressed latent space $\{z_v\}_1^V$. It simultaneously learns to decode (blue box; $D$) the latents to parameters of Gaussian splats $\mathcal{S}$, which can then be rendered back to images $x^*$.
    \textbf{Right:} We train a denoising diffusion model (pink box; $\bm{v}_\theta$) over the multi-view latent features $z_v$. This supports unconditional generation, or generation conditioned on an input image $x_\mathrm{cond}$ (itself encoded with $E$). Following the efficient, low-dimensional denoising process, the resulting latents are mapped back to a 3D scene by $D$.
    }
    \label{fig:architecture}
\end{figure}

\paragraph{Generative models.} 
Diverse families of generative models \cite{kingma2013auto, goodfellow2014generative, van2016pixel, karras2019style, DBLP:journals/corr/Sohl-DicksteinW15, ho2020denoising} have been proposed to learn complex distributions from training data. 
With the success of generative models across various modalities, including language \cite{vaswani17attention, radford2018improving}, sound \cite{oord2016wavenet}, and images \cite{Rombach_2022_CVPR, ho2022imagen}, there is now a growing interest in sampling 3D content.
A straightforward approach is to create a large-scale 3D dataset \cite{chang2015shapenet, deitke2024objaverse} and train a generative model directly on this \cite{pointclouddiffusion, vahdat2022lion, chen2023single, zhou20213d, hui2022neural, li2022diffusionsdf, cheng23sdfusion, mueller2022diffrf, bautista2022gaudi, wang2022rodin, kim2023neuralfieldldm, shue20223d, gupta20233dgen, karnewar2023holodiffusion, gu2023learning}. 
However, unlike other modalities, large-scale, highly-realistic 3D datasets of scenes are challenging to create, and most 3D representations lack shared structure (as each datapoint is created independently, e.g.~meshes have different topologies, and point-clouds different numbers of points) making learning priors difficult. 
While recent works use autodecoding \cite{bautista2022gaudi} or optimal transport to share structure across representations \cite{zhang2024gaussiancube},  
these methods have been limited to small or object-centric datasets.

To circumvent the difficulty of learning a smooth prior over independently-reconstructed scenes, 
various methods learn a 3D-aware generative model directly from images. 
A simple yet elegant approach \cite{eslami2018neural, wu2023reconfusion, blattmann2023stable, chan2023genvs, tseng2023consistent, liu2023zero1to3, Tang2023mvdiffusion, yu2023longterm, watson2022novel, kong2024eschernet, tang2024mvdiffusionpp, liu2023syncdreamer, liu2024one, hu2024mvdfusion, gao2024cat3d, tang2024lgm} is to generate multi-view images conditioned on camera poses without an explicit 3D representation, then reconstruct 3D from the generated images using classical 3D reconstruction methods. 
However, it inherits the limitations of classical methods, primarily the need to generate a large number of consistent images (>50) and the absence of priors in the 3D reconstruction process.

To directly sample 3D representations, some works learn 3D-aware generative models of 2D images, which retain the mathematical formulation of generative image models, but introduce priors into the network architecture which force the model to output images via an explicit 3D representation. 
Seminal approaches were based on VAEs \cite{kosiorek2021nerf, henderson19ijcv, henderson20cvpr, anciukevicius2022unsupervised, henderson21arxiv} and GANs \cite{schwarz2020graf, eg3d, skorokhodov2022epigraf, deng2022gram,nguyen2020blockgan, hologan, gmpi2022, devries2021unconstrained}, while current state-of-the-art methods use 3D-aware denoising diffusion models \cite{anciukevicius2022renderdiffusion, karnewar2023holodiffusion, szymanowicz2023viewset_diffusion, tewari2023forwarddiffusion, xu2023dmv3d, anciukevicius2024denoising, hoellein2024viewdiff, Schwarz2024ICLR, cao2024lightplane}. 
Unlike score-distillation methods \cite{poole2022dreamfusion, wang2024prolificdreamer, tang2023dreamgaussian, yi2023gaussiandreamer, zhou2023sparsefusion, yoo2023dreamsparse, wynn-2023-diffusionerf, zhu2023hifa, lin2023magic3d, wang2023score} that suffer from mode-seeking behavior and do not truly sample a distribution, 3D-aware diffusion models can sample 3D scenes from the true posterior distribution. 
However, existing works use radiance fields to represent the scene and thus are limited by slow training, sampling, and rendering times. 
In contrast, our work samples 3D scenes in less than a second (0.2s) compared to 5.4s for the recent \cite{szymanowicz2023viewset_diffusion} or 51s for \cite{anciukevicius2024denoising}; ours can also render the sampled 3D asset in real-time.

\section{Method}

Our goal is to build a model supporting conditional and unconditional generation of 3D scenes.
We assume access only to a training dataset of multi-view images with camera poses (readily available from phone cameras or COLMAP SfM~\cite{schoenberger2016sfm}). 
We do not require any additional 2D/3D supervision (e.g.~annotations, pretrained models, foreground masks, depth-maps) and we do not assume that camera poses are aligned consistently across the dataset.

We achieve this by designing a latent diffusion framework that is trained in two stages.
First, a 3D-aware variational autoencoder (VAE) is trained on sets of multi-view images (Sec.\ref{subsec:method-ae}). 
It encodes multi-view images to a compact latent representation, decodes this to an explicit 3D scene represented by Gaussian Splats, then renders the scene to reconstruct images.
Second, we train a denoising diffusion model on the compact latent space learnt by the autoencoder (Sec.\ref{subsec:method-denoiser}).
This diffusion model is trained jointly for class-conditional and image-conditional generation, and can efficiently learn a distribution on the latent space.
During inference, the resulting latents are decoded back to splats by the autoencoder, and rendered.

\subsection{Autoencoder}
\label{subsec:method-ae}

Our autoencoder takes as input $V$ views $\bm{x}=\{x_v\}_{v=1}^V$ of a scene (each of size $H \times W$ pixels), with their relative camera poses $\bm{\pi}=\{\pi_v\}_{v=1}^V$.
It processes these jointly to give a set of splats $\mathcal{S}$, such that rendering $\mathcal{S}$ from each $\pi_v$ should reconstruct the original image $x_v$.
Importantly, it passes all information about the scene through a low-dimensional latent bottleneck, yielding a compressed representation from which the splats are then decoded and rendered.

\paragraph{Encoding multi-view images.}
The $x_v$ are first passed independently through three downsampling residual blocks similar to \cite{esser2021taming,Rombach_2022_CVPR}, yielding feature maps of resolution $\frac{H}{8} \times \frac{W}{8}$.
These features are processed by a multi-view U-Net~\cite{RonnebergerFB15}, which enables the different views to exchange information efficiently (necessary to achieve a consistent 3D reconstruction).
This U-Net is based closely on \cite{ho2020denoising}. To adapt it to our multi-view setting, we take inspiration from video diffusion models, notably \cite{blattmann2023videoldm}, and add a small cross-view ResNet after each block that combines information from all views, for each pixel independently.
We also modify all attention layers to jointly attend across features from all views.
Aside from these parts, all the remainder of processing by residual blocks treats views independently.
The final convolution of the U-Net outputs the mean and log-variances of a feature-map of size $\frac{H}{8} \times \frac{W}{8}$ for each view.
This will be the compressed latent space $\{z_v\}_{v=1}^V$ in which we perform denoising (see Sec.\ref{subsec:method-denoiser}), and so we restrict it to have only very few channels. 
We assume a diagonal Gaussian posterior distribution, following common practice for VAEs~\cite{kingma2013auto,Rombach_2022_CVPR}.
We denote the overall encoder mapping from $\{x_v\}_{v=1}^V$ to a latent sample $\{z_v\}_{v=1}^V$ by $E$; it is shown by the green box in Fig.~\ref{fig:architecture}.

\paragraph{Decoding to a 3D scene.}
The latent features $\{z_v\}_{v=1}^V$ are decoded to a 3D scene $\mathcal{S}$ represented as Gaussian Splats~\cite{kerbl20233d}. 
Specifically, we pass the features through three upsampling residual blocks, mirroring the initial layers of $E$; this yields feature maps with the same size as the original images.
Similarly to \cite{szymanowicz2024splatter_image,charatan2023pixelsplat},
the features for each view are mapped by a convolution layer to parameters of splats supported on the view frustum.
For each pixel, we predict the depth, opacity, RGB color, rotation and scale of a corresponding splat; in total this requires 12 channels.
The 3D position of each splat is then calculated by unprojecting it along the corresponding camera ray by the predicted depth.
The union of the $V \times H \times W$ splats across all images constitutes our scene representation $\mathcal{S}$.
This representation (aptly termed a \textit{splatter image} in \cite{szymanowicz2024splatter_image}) provides a structured way to represent splats, allowing reasoning over them with standard convolutional layers instead of permutation-invariant layers necessary for unstructured point-clouds.
We denote the mapping from $\{z_v\}_{v=1}^V$ to $\mathcal{S}$ by $D$; it is shown by the blue box in Fig.~\ref{fig:architecture}.
The splats $\mathcal{S}$ can then be rendered to pixels $x^*$ using arbitrary camera parameters $\pi^*$; we denote this rendering operation by $x^*=R(\mathcal{S},\,\pi^*)$

A key benefit of having splats supported on images at all denoised viewpoints is that we can represent 3D content anywhere we look.
This contrasts with e.g.~SplatterImage~\cite{szymanowicz2024splatter_image}---when performing single-image reconstruction, they can only parameterise splats inside (or very close to) the view frustum of the input image, whereas ours can generate coherent content arbitrarily far away.

\paragraph{Conditioning on pose.}
In order to condition the autoencoder on the relative poses of the views, we design a novel strategy based on the splat renderer itself.
For each view, we generate a set of splats along the edges of its view frustum; for each view, we assign a random color.
We then render the resulting splat cloud from all cameras.
These renderings are concatenated with the input $x_v$ before the first residual block of the encoder.
Note that without this conditioning, it is impossible for the autoencoder to learn the arbitrary scene scale (even if it successfully learns to triangulate the input images), due to the perspective depth/scale ambiguity.

\paragraph{Training.}
We assume access to minibatches containing $V$ input views $\left(\bm{x}^\mathrm{(in)},\,\bm{\pi}^\mathrm{(in)}\right)$, and an additional $V'$ nearby target views $\left(\bm{x}^\mathrm{(target)},\,\bm{\pi}^\mathrm{(target)}\right)$ that are not passed to $E$.
We predict splats $\mathcal{S} = D\left(E(\bm{x}^\mathrm{(in)},\,\bm{\pi}^\mathrm{(in)})\right)$, then render these at the target views giving $x_{v'}^*=R\left(\mathcal{S},\,\pi_{v'}^\mathrm{(target)}\right) \forall v'=\{1 \ldots V'\}$.
The network is then trained as a variational autoencoder~\cite{kingma2013auto,rezende2014stochastic}, using the sum of $L^2$ and LPIPS~\cite{zhang2018unreasonable} distances between input and rendered images as a reconstruction loss, and maximising the log-probability of the sampled latents $\{z_v\}_{v=1}^V$ under a standard Gaussian prior.
This leads to the following loss:
\begin{equation}
    \mathcal{L}_\mathrm{AE} = \sum_{v'=1}^{V'} \left\{
        \big|\big|x_{v'}^* - x_{v'}^\mathrm{(target)}\big|\big|^2_2 +
        \mathrm{LPIPS}\left(x_{v'}^*,\, x_{v'}^\mathrm{(target)}\right) +
        ||z_v||^2_2
    \right\}
\end{equation}
where $\beta$ adjusts the weight of the KL loss.
Note that unlike methods using NeRFs, the speed of the splat rendering operation $R$ means we can straightforwardly apply LPIPS to full-image renderings.

\paragraph{Compression.}
Compared with treating the splats themselves as the latent space, our approach yields a compression factor of $128\times$ when we use 6 latent channels (as in our main experiments).
As shown in Sec.~\ref{sec:experiments}, this enables much more efficient training and inference for the denoiser.

\subsection{Denoiser}
\label{subsec:method-denoiser}

We now define a denoising diffusion model over the low-dimensional multi-view latent feature maps $\{z_v\}_{v=1}^V=\bm{z}$ produced by the encoder $E$.
We take a similar approach to latent diffusion models for images  \cite{Rombach_2022_CVPR}, but instead of a 2D U-Net, use a very similar multi-view U-Net architecture as in the autoencoder (Sec.~\ref{subsec:method-ae}), now conditioned on the diffusion timestep, as in \cite{ho2020denoising}.
We condition this multi-view U-Net $\hat{\bm{v}}_\theta$ on the camera poses $\pi_v$ in the same way as for the autoencoder, i.e.~concatenating a representation of all view frusta.
The U-Net is then responsible for learning the joint distribution of latent features across all views, passing information via cross-view attention and convolution operations.

\paragraph{Conditional generation.}
We train the denoiser jointly for image-conditional and class-conditional generation (choosing randomly which to use for each minibatch), and also dropping the conditioning entirely for 20\% of minibatches to enable classifier-free guidance~\cite{ho2022cfg}.
For class conditioning, we follow common practice and use a learnt embedding for class indices, which is added to the timestep embedding in the U-Net.
For image conditioning (i.e.~when performing 3D reconstruction), we again make use of our pretrained encoder $E$, to encode the conditioning images $x_\mathrm{cond}$ and their poses $\pi_\mathrm{cond}$.
The conditioning latents output by $E(x_\mathrm{cond},\,\pi_\mathrm{cond})$ are then concatenated with the noisy latents at the start of the denoiser.

\paragraph{Training.}
During training we sample minibatches of posed views $\left(\bm{x},\,\bm{\pi}\right)$. These are converted to latents $\bm{z}$ by passing them to $E$ and sampling the posterior.
We normalise $\bm{z}$ to have approximately zero mean and unit standard deviation, based on statistics of the first training minibatch.
We then sample a diffusion timestep $t$ and sample noisy latent from the Gaussian forward process $\mathcal{N}(\bm{z}^{(t)}; \alpha_t \bm{z}, \sigma_t^2 \bm{I})$, where $\alpha_t$ and $\sigma_t$ are specified by a linear noise schedule; we optimize the denoiser's parameters $\theta$ by gradient descent on the following loss:
\begin{equation}
    \mathcal{L}_\mathrm{DDM} =
    \mathbb{E}_{t,\, \bm{\epsilon}\sim\mathcal{N}(\bm{0},\bm{I}),\, \bm{z}^{(t)}} || 
    \hat{\bm{v}}_\theta(\bm{z}^{(t)}, t) - \bm{v}^{(t)})
    ||_2^2
\end{equation}
Importantly, due to our abstract latent space, our implementation trains the denoiser $\hat{\bm{v}}_\theta(\bm{z}^{(t)}, t)$ to predict $\bm{v}^{(t)} \equiv \alpha_t \bm{\epsilon} - \sigma_t \bm{z}$ \cite{salimans2022vpred} which is more numerically stable than $\bm{x}^{(0)}$ prediction used by other 3D-aware diffusion models \cite{szymanowicz2023viewset_diffusion,anciukevicius2024denoising,anciukevicius2022renderdiffusion} that constrain the denoiser itself to output rendered pixels.

\paragraph{Sampling.}
To sample a 3D scene from our model, we begin by sampling Gaussian noise $\bm{z}^{(1000)}\sim\mathcal{N}(\bm{0},\bm{I})$ in the latent space, and choosing a set of camera poses (e.g.~from a held-out validation set) as conditioning. We then use DDIM sampling~\cite{song2020denoising} to find $\bm{z}^{(0)}$, with classifier-free guidance for class conditioning.
From this we decode the generated 3D scene by $\mathcal{S} = D(\bm{z}^{(0)})$, which can then be rendered efficiently from arbitrary viewpoints $\pi^*$ using $R$.

\section{Experiments}
\label{sec:experiments}

We evaluate our model and several baselines on both generation and one-/few-view 3D reconstruction.
Further implementation details for our method and the baselines are given in the appendix.

\paragraph{Datasets.}
We evaluate our approach on two large-scale datasets of real-world images---MVImgNet~\cite{yu2023mvimgnet} and RealEstate10K~\cite{zhou2018stereomag}.
MVimgNet consists of videos showing diverse objects in the wild.
We use the splits from MVPNet subset, containing 87,820 videos covering 180 object classes, and restrict to 5000 scenes for evaluation.
Each video typically contains 30 frames.
RealEstate10K contains 69893 videos showing indoor and outdoor views of houses. We use the official splits, again limiting to 5000 scenes for evaluation.
Each video typically contains 50-200 frames. We use the complete video clips (often with substantial camera motion) for training and evaluation, not the shorter, easier segments from \cite{wiles2020synsin}.
Note that the videos in both datasets depict complete scenes, not just isolated objects.
For both datasets, we center crop the images with size equal to small edge, then rescale to $96\times96$. During training, we randomly sample sets of six frames as multi-view images.
We use the camera poses provided with each dataset, but only provide \textit{relative} poses to the model.
We do not require any canonicalisation of scene orientation or scale, unlike earlier methods \cite{szymanowicz2023viewset_diffusion,anciukevicius2022renderdiffusion,szymanowicz2024splatter_image} , and do not rely on segmentation masks nor depths.

\begin{table}[]
    \centering
    \caption{Results from our method and baselines
    on unconditional/class-conditional scene generation, 3D reconstruction from a single image, and 3D reconstruction from sparse (six) views. %
    }
    \label{tab:main-results}
    \resizebox{\linewidth}{!}{%
    \begin{small}%
    \begin{tabular}{@{} l cc ccc cc @{}}
    \toprule
         & \multicolumn{2}{c}{\textbf{Generation}} & \multicolumn{3}{c}{\textbf{1-view reconstruction}} & \multicolumn{2}{c@{}}{\textbf{6-view reconstruction}} \\
         \cmidrule{2-3}\cmidrule(l){4-6}\cmidrule(l){7-8}
         & FID $\downarrow$ & Time /s $\downarrow$ & PSNR $\uparrow$ & LPIPS $\downarrow$ & Time /s $\downarrow$ & PSNR $\uparrow$ & LPIPS $\downarrow$ \\
         \midrule
        \textbf{MVImgNet} \\
        Ours & \textbf{23.1} & \textbf{0.22} & \textbf{20.6} &  \textbf{0.324} & 0.22 & \textbf{24.7} & \textbf{0.184} \\
        SplatterImage~\cite{szymanowicz2024splatter_image} & -- & -- & 18.2 & 0.367 & \textbf{0.03} & -- & -- \\
        \midrule
        \textbf{RealEstate10K} \\
        Ours & \textbf{29.5} & \textbf{0.22} & \textbf{16.4} & 0.455 & 0.22 & \textbf{23.3} & \textbf{0.155} \\
        SplatterImage~\cite{szymanowicz2024splatter_image} & -- & -- & 16.2 & \textbf{0.347} & \textbf{0.03} & -- & -- \\
        \midrule
        \textbf{MVImgNet furniture} \\
        Ours & \textbf{89.0} & \textbf{0.22} & 17.9 & \textbf{0.407} & 0.22 & 22.9 & 0.214 \\
        GIBR~\cite{anciukevicius2024denoising} & 99.8 & 44.9 & \textbf{18.5} & 0.414 & 44.3 & \textbf{25.4} & \textbf{0.199} \\
        Viewset Diffusion~\cite{szymanowicz2023viewset_diffusion} & 191.4 & 4.98 & 17.6 & 0.540 & 4.25 & -- & -- \\
        RenderDiffusion~\cite{anciukevicius2022renderdiffusion,anciukevicius2024denoising} & 234.1 & 10.2 & 17.4 & 0.622 & 10.2 & 18.4 & 0.601 \\
        PixelNeRF~\cite{yu2021pixelnerf,anciukevicius2024denoising} & -- & -- & 16.6 & 0.582 & \textbf{0.12} & 15.7 & 0.647 \\
        \bottomrule
    \end{tabular}%
    \end{small}%
}
\end{table}

\paragraph{Baselines.}
We compare our approach to several existing works.
\textbf{GIBR}~\cite{anciukevicius2024denoising} and \textbf{ViewSet Diffusion}~\cite{szymanowicz2023viewset_diffusion} are 3D-aware diffusion models over multi-view images. They encode sets of noisy views, and uses these to build a radiance field representation of the scene that is then rendered to give the denoised images. Like our method, they support both unconditional generation and reconstruction.
However, both are relatively expensive since they must perform volumetric rendering during every denoising step.
\textbf{RenderDiffusion}~\cite{anciukevicius2022renderdiffusion} is a similar method that only requires single images during training; we use the variant adapted for the in-the-wild data by \cite{anciukevicius2024denoising}.
\textbf{SplatterImage}~\cite{szymanowicz2024splatter_image} is a recent deterministic method for 3D reconstruction from a single image, outputting splats from a single U-Net pass.
This is currently the only method able to directly predict 3D splats from a single image.
However, the original work only considers masked views of isolated objects; we therefore adapt their method to work for our larger, unmasked scenes.
Lastly, \textbf{PixelNeRF}~\cite{yu2021pixelnerf} predicts a radiance field deterministically by unprojecting features from one or more input images; we use the in-the-wild variant from \cite{anciukevicius2024denoising}.

\begin{figure}[htbp]
\centering
    \begin{subfigure}{\textwidth}
        \resizebox{\textwidth}{!}{
            \begin{tikzpicture}
              \node[inner sep=0] (image1) {\includegraphics[height=5cm]{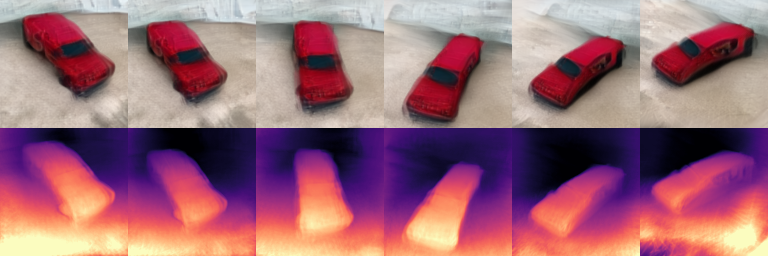}};
              \node[rotate=90, anchor=south, font=\LARGE] at (image1.west) {\texttt{\textcolor{GHOST}{pl}car\textcolor{GHOST}{pl}}};  
            \end{tikzpicture}
            \begin{tikzpicture}
              \node[inner sep=0] (image2) {\includegraphics[height=5cm]{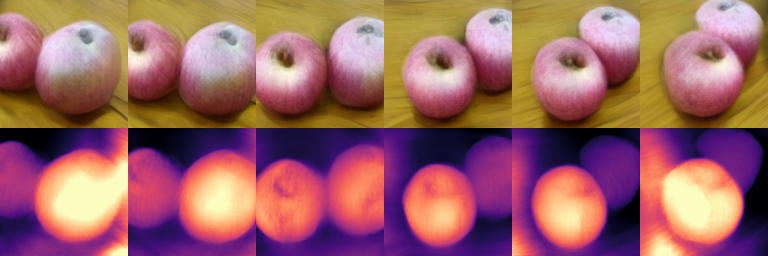}};
              \node[rotate=90, anchor=south, font=\LARGE] at (image2.west) {\texttt{apple}};  
            \end{tikzpicture}
            \begin{tikzpicture}
              \node[inner sep=0] (image3) {\includegraphics[height=5cm]{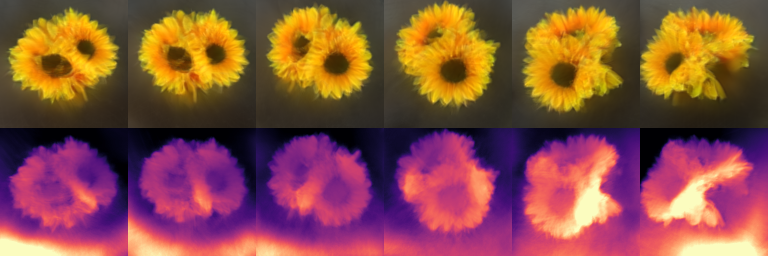}};
              \node[rotate=90, anchor=south, font=\LARGE] at (image3.west) {\texttt{\textcolor{GHOST}{p}sunflower\textcolor{GHOST}{p}}};  
            \end{tikzpicture}
        }
        \resizebox{\textwidth}{!}{
            \begin{tikzpicture}
              \node[inner sep=0] (image1) {\includegraphics[height=5cm]{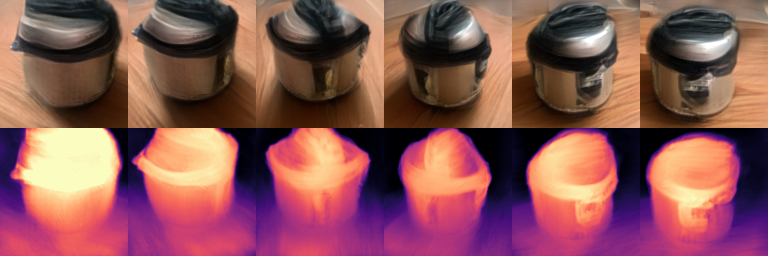}};
              \node[rotate=90, anchor=south, font=\LARGE] at (image1.west) {\texttt{pressure cooker}};  
            \end{tikzpicture}
            \begin{tikzpicture}
              \node[inner sep=0] (image2) {\includegraphics[height=5cm]{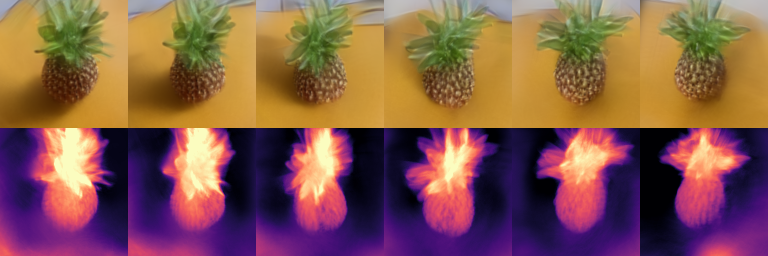}};
              \node[rotate=90, anchor=south, font=\LARGE] at (image2.west) {\texttt{pineapple}};  
            \end{tikzpicture}
            \begin{tikzpicture}
              \node[inner sep=0] (image3) {\includegraphics[height=5cm]{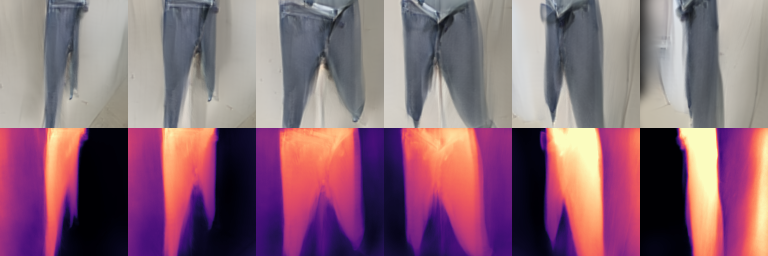}};
              \node[rotate=90, anchor=south, font=\LARGE] at (image3.west) {\texttt{\textcolor{GHOST}{l}pants\textcolor{GHOST}{l}}};  
            \end{tikzpicture}%
        }
    \caption{MVImgNet}
    \label{subfig:gen-mvimgnet}
    \end{subfigure}
    
    \vspace{4pt}
    
    \begin{subfigure}{\textwidth}
        \resizebox{\textwidth}{!}{%
            \begin{tikzpicture}
              \node[inner sep=0] (image1) {\includegraphics[height=5cm]{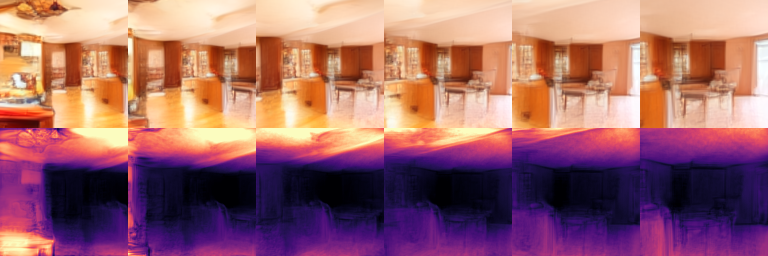}};
            \end{tikzpicture}
            \hspace{1em}
            \begin{tikzpicture}
              \node[inner sep=0] (image2) {\includegraphics[height=5cm]{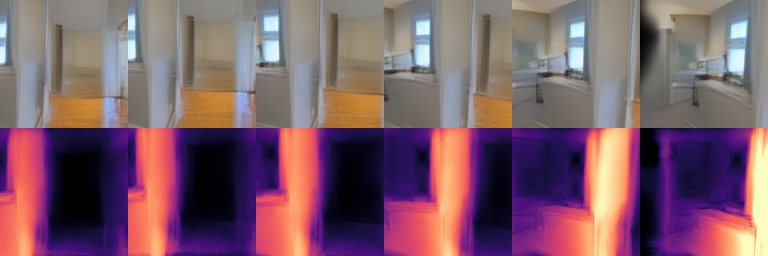}};
            \end{tikzpicture}%
        }
    \caption{RealEstate10K}
    \label{subfig:gen-re10k}
    \end{subfigure}
    \caption{Qualitative examples of class-conditional (MVImgNet) and unconditional 3D generations (RealEstate10K) from our method. For each example, the top row shows six rendered views of the sampled 3D scene, while the bottom row shows the corresponding depths. Note that our model samples 3D scenes containing objects with complex shape on a realistic background. }
    \label{fig:results-gen}
\end{figure}
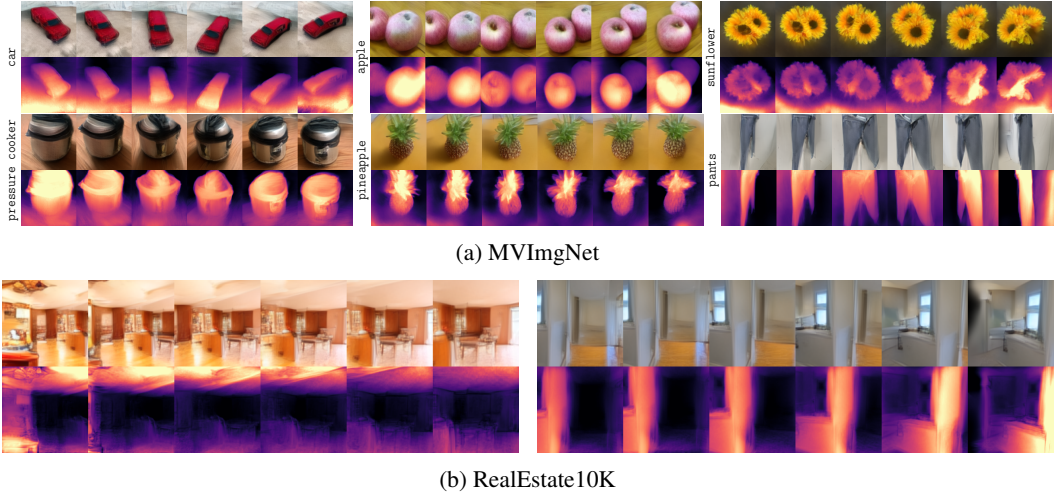

\paragraph{Generation results.}
We first evaluate on unconditional generation of 3D scenes (for RealEstate10K), and class-conditional generation for the full (180 classes) MVImgNet dataset.
Qualitative examples are given in Fig.~\ref{fig:results-gen}, with more examples in the appendix.
We see that our model can generate objects of diverse classes, given just the label as conditioning. The generated scenes are coherent across views, including during long camera motions in RealEstate10K.
We also perform a quantitative evaluation against several existing methods.
Here we closely follow the setting of \cite{anciukevicius2024denoising}, using the \textit{chair}, \textit{table} and \textit{sofa} classes of MVImgNet.
Specifically, we compare against GIBR~\cite{anciukevicius2024denoising}, ViewSet Diffusion~\cite{szymanowicz2023viewset_diffusion} and also the earlier RenderDiffusion~\cite{anciukevicius2022renderdiffusion}.
The results are presented in Table~\ref{tab:main-results} (bottom five rows).
On class-conditional generation, our method significantly out-performs the baselines, achieving an FID of 89.0, vs 99.8 for the second best (GIBR).
Moreover, it achieves this while being $174\times$ faster than GIBR (just 0.22s for ours, vs 45s for GIBR), since it avoids the need to render an image at every denoising step, and also operates over a lower-dimensionality space.

\begin{figure}[t]
\centering
    \begin{subfigure}{\textwidth}
    \resizebox{\textwidth}{!}{%
            \begin{minipage}[c]{0.05\textwidth}
                \resizebox{\textwidth}{!}{%
                    \begin{tikzpicture}
                      \node[inner sep=0] (image1) {\includegraphics[height=2cm]{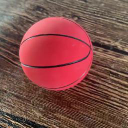}};
                    \end{tikzpicture}
                }
            \end{minipage}
            \begin{minipage}[c]{0.25\textwidth}
                \resizebox{\textwidth}{!}{%
                    \begin{tikzpicture}
                      \node[inner sep=0] (image1) {\includegraphics[height=2cm, clip, trim=9.2cm 0 54.4cm 0]{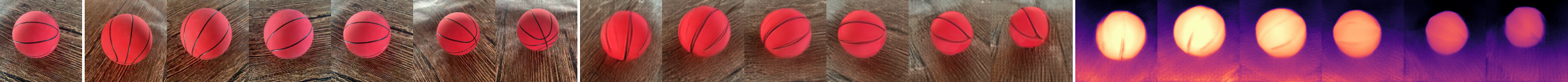}};
                    \end{tikzpicture}
                }
            \end{minipage}
            \begin{minipage}[c]{0.55\textwidth}
                \resizebox{\textwidth}{!}{%
                    \begin{tikzpicture}
                      \node[inner sep=0] (image1) {\includegraphics[height=2cm, clip, trim=31.8cm 0 0 0]{figures/1-im-recon/mvimgnet/ours/2a002f7d_02.png}};

                    \end{tikzpicture}
                }
                \vspace{0.1em}
                \resizebox{\textwidth}{!}{%
                    \begin{tikzpicture}
                      \node[inner sep=0] (image2) {\includegraphics[height=2cm, clip, trim=23.9cm 0 0 0]{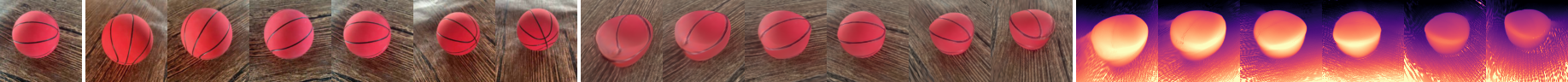}};
                    \end{tikzpicture}
                }
            \end{minipage}
    }
    \resizebox{\textwidth}{!}{%
            \begin{minipage}[c]{0.05\textwidth}
                \resizebox{\textwidth}{!}{%
                    \begin{tikzpicture}
                      \node[inner sep=0] (image1) {\includegraphics[height=2cm]{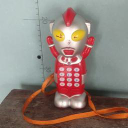}};
                    \end{tikzpicture}
                }
            \end{minipage}
            \begin{minipage}[c]{0.25\textwidth}
                \resizebox{\textwidth}{!}{%
                    \begin{tikzpicture}
                      \node[inner sep=0] (image1) {\includegraphics[height=2cm, clip, trim=9.2cm 0 54.4cm 0]{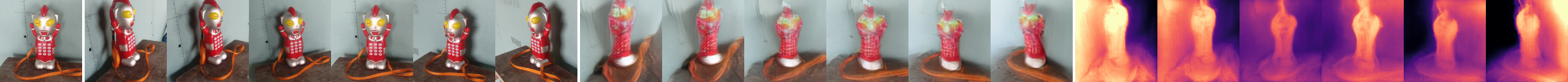}};
                    \end{tikzpicture}
                }
            \end{minipage}
            \begin{minipage}[c]{0.55\textwidth}
                \resizebox{\textwidth}{!}{%
                    \begin{tikzpicture}
                      \node[inner sep=0] (image1) {\includegraphics[height=2cm, clip, trim=31.8cm 0 0 0]{figures/1-im-recon/mvimgnet/ours/260052d1_00.png}};

                    \end{tikzpicture}
                }
                \vspace{0.1em}
                \resizebox{\textwidth}{!}{%
                    \begin{tikzpicture}
                      \node[inner sep=0] (image2) {\includegraphics[height=2cm, clip, trim=23.9cm 0 0 0]{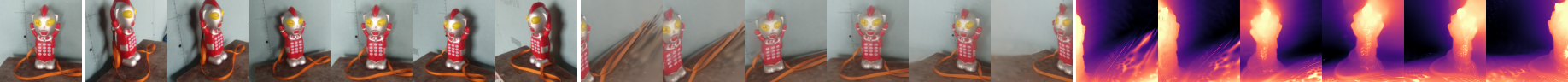}};
                    \end{tikzpicture}
                }
            \end{minipage}
    }
    \resizebox{\textwidth}{!}{%
            \begin{minipage}[c]{0.05\textwidth}
                \resizebox{\textwidth}{!}{%
                    \begin{tikzpicture}
                      \node[inner sep=0] (image1) {\includegraphics[height=2cm]{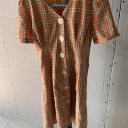}};
                    \end{tikzpicture}
                }
            \end{minipage}
            \begin{minipage}[c]{0.25\textwidth}
                \resizebox{\textwidth}{!}{%
                    \begin{tikzpicture}
                      \node[inner sep=0] (image1) {\includegraphics[height=2cm, clip, trim=9.2cm 0 54.4cm 0]{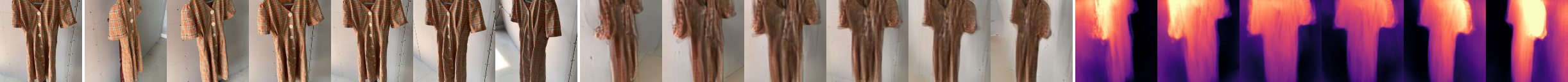}};
                    \end{tikzpicture}
                }
            \end{minipage}
            \begin{minipage}[c]{0.55\textwidth}
                \resizebox{\textwidth}{!}{%
                    \begin{tikzpicture}
                      \node[inner sep=0] (image1) {\includegraphics[height=2cm, clip, trim=31.8cm 0 0 0]{figures/1-im-recon/mvimgnet/ours/21005ead_00.png}};

                    \end{tikzpicture}
                }
                \vspace{0.1em}
                \resizebox{\textwidth}{!}{%
                    \begin{tikzpicture}
                      \node[inner sep=0] (image2) {\includegraphics[height=2cm, clip, trim=23.9cm 0 0 0]{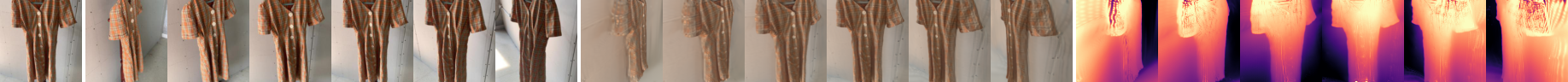}};
                    \end{tikzpicture}
                }
            \end{minipage}
    }
    \resizebox{\textwidth}{!}{%
            \begin{minipage}[c]{0.05\textwidth}
                \resizebox{\textwidth}{!}{%
                    \begin{tikzpicture}
                      \node[inner sep=0] (image1) {\includegraphics[height=2cm]{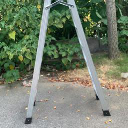}};
                    \end{tikzpicture}
                }
            \end{minipage}
            \begin{minipage}[c]{0.25\textwidth}
                \resizebox{\textwidth}{!}{%
                    \begin{tikzpicture}
                      \node[inner sep=0] (image1) {\includegraphics[height=2cm, clip, trim=9.2cm 0 54.4cm 0]{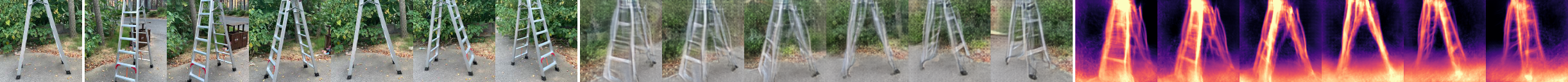}};
                    \end{tikzpicture}
                }
            \end{minipage}
            \begin{minipage}[c]{0.55\textwidth}
                \resizebox{\textwidth}{!}{%
                    \begin{tikzpicture}
                      \node[inner sep=0] (image1) {\includegraphics[height=2cm, clip, trim=31.8cm 0 0 0]{figures/1-im-recon/mvimgnet/ours/030083be_02.png}};

                    \end{tikzpicture}
                }
                \vspace{0.1em}
                \resizebox{\textwidth}{!}{%
                    \begin{tikzpicture}
                      \node[inner sep=0] (image2) {\includegraphics[height=2cm, clip, trim=23.9cm 0 0 0]{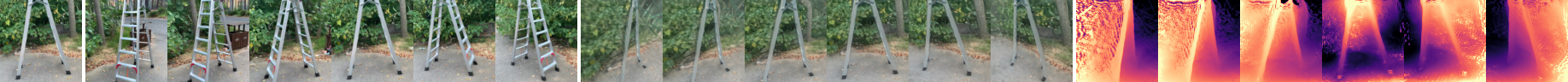}};
                    \end{tikzpicture}
                }
            \end{minipage}
    }
    \caption{MVImgNet}
    \label{subfig:1-im-recon-mvimgnet}
    \end{subfigure}

    \vspace{4pt}

    \begin{subfigure}{\textwidth}
    \resizebox{\textwidth}{!}{%
            \begin{minipage}[c]{0.05\textwidth}
                \resizebox{\textwidth}{!}{%
                    \begin{tikzpicture}
                      \node[inner sep=0] (image1) {\includegraphics[height=2cm]{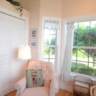}};
                    \end{tikzpicture}
                }
            \end{minipage}
            \begin{minipage}[c]{0.25\textwidth}
                \resizebox{\textwidth}{!}{%
                    \begin{tikzpicture}
                      \node[inner sep=0] (image1) {\includegraphics[height=2cm, clip, trim=9.2cm 0 54.4cm 0]{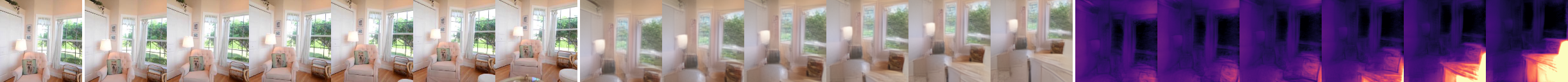}};
                    \end{tikzpicture}
                }
            \end{minipage}
            \begin{minipage}[c]{0.55\textwidth}
                \resizebox{\textwidth}{!}{%
                    \begin{tikzpicture}
                      \node[inner sep=0] (image1) {\includegraphics[height=2cm, clip, trim=31.8cm 0 0 0]{figures/1-im-recon/ra10k/ours/004dd4b46a06e5be_02.png}};

                    \end{tikzpicture}
                }
                \vspace{0.1em}
                \resizebox{\textwidth}{!}{%
                    \begin{tikzpicture}
                      \node[inner sep=0] (image2) {\includegraphics[height=2cm, clip, trim=23.9cm 0 0 0]{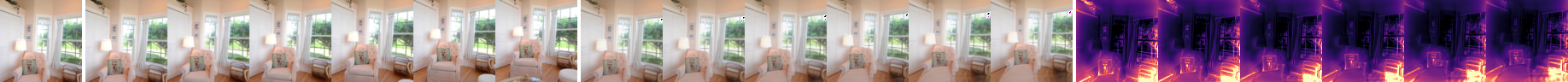}};
                    \end{tikzpicture}
                }
            \end{minipage}
    }
    \resizebox{\textwidth}{!}{%
            \begin{minipage}[c]{0.05\textwidth}
                \resizebox{\textwidth}{!}{%
                    \begin{tikzpicture}
                      \node[inner sep=0] (image1) {\includegraphics[height=2cm]{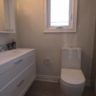}};
                    \end{tikzpicture}
                }
            \end{minipage}
            \begin{minipage}[c]{0.25\textwidth}
                \resizebox{\textwidth}{!}{%
                    \begin{tikzpicture}
                      \node[inner sep=0] (image1) {\includegraphics[height=2cm, clip, trim=9.2cm 0 54.4cm 0]{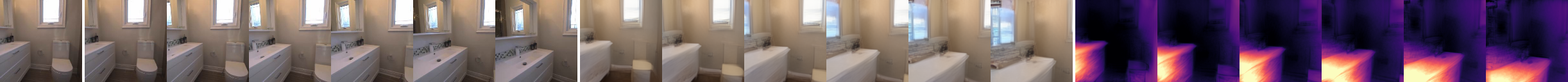}};
                    \end{tikzpicture}
                }
            \end{minipage}
            \begin{minipage}[c]{0.55\textwidth}
                \resizebox{\textwidth}{!}{%
                    \begin{tikzpicture}
                      \node[inner sep=0] (image1) {\includegraphics[height=2cm, clip, trim=31.8cm 0 0 0]{figures/1-im-recon/ra10k/ours/03ac738af49c7596_03.png}};

                    \end{tikzpicture}
                }
                \vspace{0.1em}
                \resizebox{\textwidth}{!}{%
                    \begin{tikzpicture}
                      \node[inner sep=0] (image2) {\includegraphics[height=2cm, clip, trim=23.9cm 0 0 0]{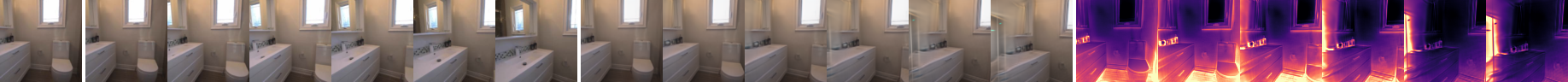}};
                    \end{tikzpicture}
                }
            \end{minipage}
    }
    \caption{RealEstate10k}
    \label{subfig:1-im-recon-ra10k}
    \end{subfigure}
 \caption{Qualitative comparison of 3D reconstruction from a single image between our model (top row of each scene) and SplatterImage \cite{szymanowicz2024splatter_image} (bottom row of each scene) on MVImgNet (a) and RealEstate10k (b). The first column shows the input (conditioning) image, the second displays the ground truth images, while the third and fourth columns display the predicted frames and depths, respectively. Compared to the baseline, our model yields more plausible reconstruction, especially of the occluded regions and the background. Additional examples in the Appendix.}
    \label{fig:results-1-im-recon}
\end{figure}

\begin{figure}[t]
    \centering
    \resizebox{0.7\textwidth}{!}{%
            \begin{minipage}[c]{0.12\textwidth}
                \resizebox{\textwidth}{!}{%
                    \begin{tikzpicture}
                      \node[inner sep=0] (image1) {\includegraphics[height=2cm, clip, trim=0 0 4.6cm  0]{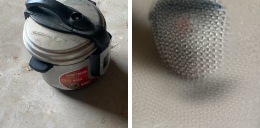}};
                    \end{tikzpicture}
                }
            \end{minipage}
            \begin{minipage}[c]{0.68\textwidth}
                \resizebox{\textwidth}{!}{%
                    \begin{tikzpicture}
                      \node[inner sep=0] (image1) {\includegraphics[height=2cm, clip, trim=13.7cm 0 9cm  0]{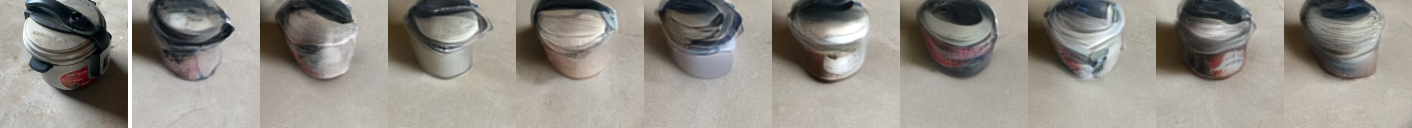}};
                    \end{tikzpicture}
                }
            \end{minipage}
            \begin{minipage}[c]{0.12\textwidth}
                \resizebox{\textwidth}{!}{%
                    \begin{tikzpicture}
                      \node[inner sep=0] (image1) {\includegraphics[height=2cm, clip, trim=4.6cm 0 0 0]{figures/back-diversity/det/1800ce8b_backs.png}};
                    \end{tikzpicture}
                }
            \end{minipage}
    }
    \resizebox{0.7\textwidth}{!}{%
            \begin{minipage}[c]{0.12\textwidth}
                \resizebox{\textwidth}{!}{%
                    \begin{tikzpicture}
                      \node[inner sep=0] (image1) {\includegraphics[height=2cm, clip, trim=0 0 4.6cm  0]{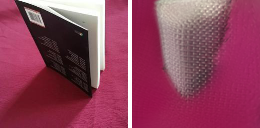}};
                    \end{tikzpicture}
                }
            \end{minipage}
            \begin{minipage}[c]{0.68\textwidth}
                \resizebox{\textwidth}{!}{%
                    \begin{tikzpicture}
                      \node[inner sep=0] (image1) {\includegraphics[height=2cm, clip, trim=13.7cm 0 9cm  0]{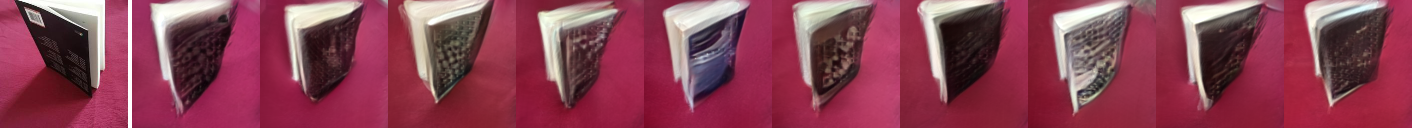}};
                    \end{tikzpicture}
                }
            \end{minipage}
            \begin{minipage}[c]{0.12\textwidth}
                \resizebox{\textwidth}{!}{%
                    \begin{tikzpicture}
                      \node[inner sep=0] (image1) {\includegraphics[height=2cm, clip, trim=4.6cm 0 0 0]{figures/back-diversity/det/1c012ff3_backs.png}};
                    \end{tikzpicture}
                }
            \end{minipage}       
    }
    \caption{Given a single input image from MVImgNet (first column), our model performs 3D reconstruction in a generative manner, and can therefore produce multiple diverse back-views (columns 2 through 7). When compared with the back-view generated by a deterministic model (column 8), our model's predictions are much sharper. Additional examples in the Appendix.}
    \label{fig:results-diverse-backs}
\end{figure}

\paragraph{Single-view 3D reconstruction results.}
We now evaluate generative 3D reconstruction from a single image. Given an image, we predict 11 other evenly-spaced viewpoints, and measure the accuracy of the predicted views using PSNR and LPIPS~\cite{zhang2018unreasonable}.
Since our method is generative (and can generate many plausible samples for a given input), we follow standard practice for stochastic prediction and draw multiple (20) samples for each scene then record the best.
Here we compare against SplatterImage~\cite{szymanowicz2024splatter_image} on MVImgNet and RealEstate10K, and all other baselines on the furniture subset of MVImgNet.
Qualitative results are given in Fig.~\ref{fig:results-1-im-recon}.
We see that our method can reconstruct plausible shapes from a single image, for diverse object classes and even entire rooms. Details in the input image are preserved, while plausible content is generated in occluded parts.
Moreover, in Fig.~\ref{fig:results-diverse-backs}, we show that given one image of an object, our model can sample diverse (yet plausible) texture and shape for the unobserved back of the object.
Quantitative results are presented in Table~\ref{tab:main-results} (`1-view reconstruction' columns).
We see that our method performs significantly better than SplatterImage according to both PSNR and LPIPS on the full MVImgNet dataset, as well as PSNR on RealEstate10K, though SplatterImage slightly exceeds it according to LPIPS.
On the MVImgNet furniture subset (following the protocol of \cite{anciukevicius2024denoising}, our method is best with respect to the perceptual LPIPS metric, while GIBR performs slightly better according to PSNR (which generally favors blurrier results).
The older, deterministic PixelNeRF method performs worst according to both reconstruction metrics.
In the appendix, we include additional quantitative results showing that our method's quality and accuracy are comparable for images rendered from the same viewpoints as supporting the splats and latents, and images rendered from other held-out viewpoints.

We also measure the time to reconstruct a scene using each of these methods (fixing each to 50 denoising steps for fairness), processing a minibatch of 8 scenes then calculating the average time per scene, on a single consumer GPU (NVIDIA RTX 3090).
Of the generative methods, ours is by far the fastest (0.22s), compared with 4.3s for the next quickest (ViewSet Diffusion). GIBR is the slowest, taking 44s to reconstruct a scene; it is limited by the need to construct and re-render a NeRF during each denoising step.
PixelNeRF is fastest overall, but trades off quality for speed and is unable to represent a posterior over 3D scenes.

\paragraph{Sparse-view 3D reconstruction results.}
Lastly, we evaluate reconstruction from sparse (6) views, measuring PSNR and LPIPS at six held-out views spaced equally between the inputs.
Quantitative results are given in Table~\ref{tab:main-results} (`6-view reconstruction' columns), while qualitative examples are presented in the appendix (Fig.~\ref{fig:6-im-recon-app}).
We see that qualitative reconstruction performance is very good, with fine details preserved, and plausible depth-maps.
Quantitatively, our method slightly under-performs GIBR on the furniture subset of MVImgNet, although it executes much faster.
Note that for this task we use only the autoencoder of our model, not the denoiser.
Thus, these results also demonstrate that the autoencoder faithfully preserves details in 3D scenes, even while compressing them by $128\times$, justifying its use as a latent space over which to learn a generative prior.

\begin{figure}[t]
    \centering
    \includegraphics[width=\linewidth]{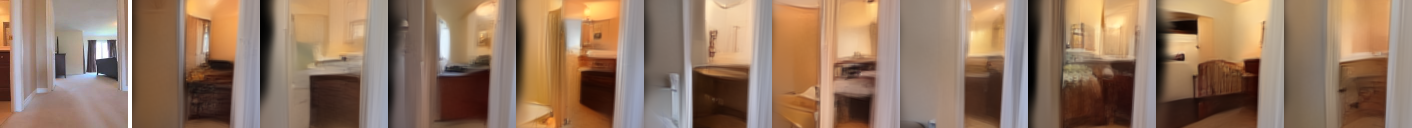}
    \includegraphics[width=\linewidth]{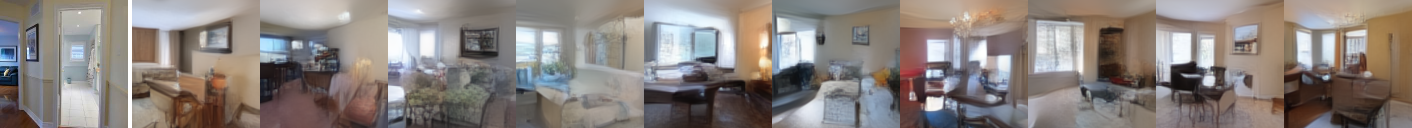}
    \caption{Given a single input image from RealEstate10K (first column), our model generates diverse possible completions of parts of the house that are not initially visible. In the top row, the camera moves into to the doorway to the left; in the bottom row, the camera moves along the hallway. In both cases, our model generates diverse samples for the room that is revealed}
    \label{fig:results-diverse-backs-re10k}
\end{figure}

\begin{table}
    \centering
    \caption{Non-generative and non-latent ablations of our method, on MVImgNet. Note we train all models with an equal compute budget.}
    \label{tab:non-gen-non-latent}
    \begin{tabular}{@{} l cccc @{}}
    \toprule
      & {FID $\downarrow$} & {PSNR $\uparrow$} & {LPIPS $\downarrow$} & {Time /s $\downarrow$} \\
      \midrule
      Full model   & \textbf{23.1} & \textbf{20.6} & \textbf{0.324} & 0.22  \\
      Deterministic & -- & 20.0 & 0.466 & \textbf{0.02} \\
      Splats-as-latents  & 57.3 & 19.2 & 0.389 & 13.9 \\
      Non-latent  & 133.8 & 18.9 & 0.499 & 4.88 \\
      \bottomrule
    \end{tabular}
\end{table}

\paragraph{Benefit of being generative and latent.}
We now evaluate several variants of our model, demonstrating the benefit of operating on latent space, and of treating 3D reconstruction as a generative (probabilistic) task, rather than deterministic.
Results on MVImgNet are presented in Table~\ref{tab:non-gen-non-latent}.
Here \textbf{Deterministic} denotes a variant of our model where the denoiser is replaced by a deterministic predictor of the latent variables representing the scene, given one input image. This model has the same network architecture as the main model, but does not sample a posterior distribution over possible scenes; instead it is expected to learn the conditional expectation in latent space.
We see this scores substantially lower than our model on LPIPS, and slightly lower on PSNR. This is expected since PSNR (which is based on mean square error) penalises overly-smoothed solutions less heavily than LPIPS (which prefers perceptually-correct outputs, e.g.~sharp edges even if these are slightly misaligned).
Moreover, qualitative results in Fig.~\ref{fig:results-diverse-backs} show that where our method generates many sharp, plausible samples for unobserved parts of the scene, the deterministic variant yields a blurry prediction that averages away the uncertainty.
Moreover, for RealEstate10K (Fig.~\ref{fig:results-diverse-backs-re10k}), our model can generate diverse yet plausible contents for rooms that are not visible in the input frame, e.g.~when the camera moves through a doorway.
\textbf{Splats-as-latents} denotes a natural ablation where instead of learning a compressed latent space, we instead treat the multi-view splat parameters themselves as the latent variables, and learn a denoiser directly over these.
We train this ablation for the same duration as our main model, to fairly measure the trade-off of performance vs accuracy.
We see that both generation and reconstruction performance is significantly worse than ours (57.3 vs 23.1 FID, and 0.39 vs 0.32 LPIPS). Moreover, due to the $128\times$ larger dimensionality, this model takes more than $60\times$ longer than ours to sample a 3D scene.
Thus, for a fixed training compute budget, our latent approach yields a clear benefit in terms of quality and accuracy.
Lastly, \textbf{Non-latent} is an ablation that uses a single-stage training process, similar to \cite{anciukevicius2024denoising,szymanowicz2023viewset_diffusion}. Here the denoiser operates directly over pixels, but incorporates our splat-based scene representation within the decoder, with clean pixels given by rendering this at each denoising step.
This model performs worse still (for equal training compute budget), reaching an FID of 133.8 and LPIPS of 0.499.

\section{Conclusion}
\label{sec:conclusion}

\paragraph{Limitations.}
While using view-supported splats enables fast optimisation and real-time rendering, our model must learn to align splats supported on different views.
This sometimes leads to artifacts at the edges of objects, where splats associated with different views are slightly misaligned.
Future work could explore using a hybrid 2D-3D architecture to reason over splats directly in 3D space.
Also, though our model trains on in-the-wild datasets of multi-view images, we still assume that each scene is static, which prevents training on arbitrary videos.
Lastly, we still require camera poses for each image; while this is easily satisfied, it would be preferable to remove this requirement.

\paragraph{Conclusion.}
We introduced a latent diffusion model that samples and reconstructs large real-world scenes represented as 3D Gaussians in as little as 0.2 seconds. 
Our evaluations showed that this model performs $20\times$ faster than the most recent 3D-aware diffusion models whilst maintaining scene realism and enabling real-time rendering.
Trained solely on posed multi-view images, without explicit supervision, our model enables the use of diverse, in-the-wild datasets and paves the way for broad adoption of generative models in the 3D domain.

\begin{ack}

PH was supported in part by the Royal Society (RGS\textbackslash{}R2\textbackslash{}222045). DI was supported by EPSRC (EP/R513222/1). TA was supported in part by an EPSRC Doctoral Training Partnership.

\end{ack}

{
\small
\bibliographystyle{abbrv}
\bibliography{references}

\begin{thebibliography}{100}

\bibitem{anciukevicius2022unsupervised}
T.~Anciukevicius, P.~Fox-Roberts, E.~Rosten, and P.~Henderson.
\newblock Unsupervised causal generative understanding of images.
\newblock {\em Advances in Neural Information Processing Systems},
  35:37037--37054, 2022.

\bibitem{anciukevicius2024denoising}
T.~Anciukevi{\v{c}}ius, F.~Manhardt, F.~Tombari, and P.~Henderson.
\newblock Denoising diffusion via image-based rendering.
\newblock In {\em The Twelfth International Conference on Learning
  Representations}, 2024.

\bibitem{anciukevicius2022renderdiffusion}
T.~Anciukevi{\v{c}}ius, Z.~Xu, M.~Fisher, P.~Henderson, H.~Bilen, N.~J. Mitra,
  and P.~Guerrero.
\newblock Renderdiffusion: Image diffusion for 3d reconstruction, inpainting
  and generation.
\newblock In {\em Proceedings of the IEEE/CVF Conference on Computer Vision and
  Pattern Recognition (CVPR)}, pages 12608--12618, June 2023.

\bibitem{armeni2017joint}
I.~Armeni, S.~Sax, A.~R. Zamir, and S.~Savarese.
\newblock Joint 2d-3d-semantic data for indoor scene understanding.
\newblock {\em arXiv preprint arXiv:1702.01105}, 2017.

\bibitem{barron2021mip}
J.~T. Barron, B.~Mildenhall, M.~Tancik, P.~Hedman, R.~Martin-Brualla, and P.~P.
  Srinivasan.
\newblock Mip-nerf: A multiscale representation for anti-aliasing neural
  radiance fields.
\newblock In {\em Proceedings of the IEEE/CVF International Conference on
  Computer Vision}, pages 5855--5864, 2021.

\bibitem{bautista2022gaudi}
M.~A. Bautista, P.~Guo, S.~Abnar, W.~Talbott, A.~Toshev, Z.~Chen, L.~Dinh,
  S.~Zhai, H.~Goh, D.~Ulbricht, A.~Dehghan, and J.~Susskind.
\newblock Gaudi: A neural architect for immersive 3d scene generation.
\newblock In {\em NeurIPS}, 2022.

\bibitem{blattmann2023stable}
A.~Blattmann, T.~Dockhorn, S.~Kulal, D.~Mendelevitch, M.~Kilian, D.~Lorenz,
  Y.~Levi, Z.~English, V.~Voleti, A.~Letts, et~al.
\newblock Stable video diffusion: Scaling latent video diffusion models to
  large datasets.
\newblock {\em arXiv preprint arXiv:2311.15127}, 2023.

\bibitem{blattmann2023videoldm}
A.~Blattmann, R.~Rombach, H.~Ling, T.~Dockhorn, S.~W. Kim, S.~Fidler, and
  K.~Kreis.
\newblock Align your latents: High-resolution video synthesis with latent
  diffusion models.
\newblock In {\em IEEE Conference on Computer Vision and Pattern Recognition
  ({CVPR})}, 2023.

\bibitem{cao2024lightplane}
A.~Cao, J.~Johnson, A.~Vedaldi, and D.~Novotny.
\newblock Lightplane: Highly-scalable components for neural 3d fields.
\newblock {\em ArXiv}, 2024.

\bibitem{eg3d}
E.~R. Chan, C.~Z. Lin, M.~A. Chan, K.~Nagano, B.~Pan, S.~D. Mello, O.~Gallo,
  L.~Guibas, J.~Tremblay, S.~Khamis, T.~Karras, and G.~Wetzstein.
\newblock Efficient geometry-aware {3D} generative adversarial networks.
\newblock In {\em CVPR}, 2022.

\bibitem{chan2023genvs}
E.~R. Chan, K.~Nagano, J.~J. Park, M.~Chan, A.~W. Bergman, A.~Levy, M.~Aittala,
  S.~D. Mello, T.~Karras, and G.~Wetzstein.
\newblock Generative novel view synthesis with 3d-aware diffusion models.
\newblock In {\em IEEE International Conference on Computer Vision (ICCV)},
  October 2023.

\bibitem{chang2015shapenet}
A.~X. Chang, T.~Funkhouser, L.~Guibas, P.~Hanrahan, Q.~Huang, Z.~Li,
  S.~Savarese, M.~Savva, S.~Song, H.~Su, et~al.
\newblock Shapenet: An information-rich 3d model repository.
\newblock {\em arXiv preprint arXiv:1512.03012}, 2015.

\bibitem{charatan2023pixelsplat}
D.~Charatan, S.~Li, A.~Tagliasacchi, and V.~Sitzmann.
\newblock pixelsplat: 3d gaussian splats from image pairs for scalable
  generalizable 3d reconstruction.
\newblock {\em arXiv preprint arXiv:2312.12337}, 2023.

\bibitem{chen2022tensorf}
A.~Chen, Z.~Xu, A.~Geiger, J.~Yu, and H.~Su.
\newblock Tensorf: Tensorial radiance fields.
\newblock In {\em Computer Vision – ECCV 2022: 17th European Conference, Tel
  Aviv, Israel, October 23–27, 2022, Proceedings, Part XXXII}, page
  333–350, Berlin, Heidelberg, 2022. Springer-Verlag.

\bibitem{mvsnerf}
A.~Chen, Z.~Xu, F.~Zhao, X.~Zhang, F.~Xiang, J.~Yu, and H.~Su.
\newblock Mvsnerf: Fast generalizable radiance field reconstruction from
  multi-view stereo.
\newblock In {\em Proceedings of the IEEE/CVF International Conference on
  Computer Vision}, pages 14124--14133, 2021.

\bibitem{chen2023single}
H.~Chen, J.~Gu, A.~Chen, W.~Tian, Z.~Tu, L.~Liu, and H.~Su.
\newblock Single-stage diffusion nerf: A unified approach to 3d generation and
  reconstruction.
\newblock In {\em ICCV}, 2023.

\bibitem{chen2024mvsplat}
Y.~Chen, H.~Xu, C.~Zheng, B.~Zhuang, M.~Pollefeys, A.~Geiger, T.-J. Cham, and
  J.~Cai.
\newblock Mvsplat: Efficient 3d gaussian splatting from sparse multi-view
  images.
\newblock {\em arXiv preprint arXiv:2403.14627}, 2024.

\bibitem{cheng23sdfusion}
Y.-C. Cheng, H.-Y. Lee, S.~Tuyakov, A.~Schwing, and L.~Gui.
\newblock {SDFusion}: Multimodal 3d shape completion, reconstruction, and
  generation.
\newblock In {\em CVPR}, 2023.

\bibitem{chung2023luciddreamer}
J.~Chung, S.~Lee, H.~Nam, J.~Lee, and K.~M. Lee.
\newblock Luciddreamer: Domain-free generation of 3d gaussian splatting scenes.
\newblock {\em arXiv preprint arXiv:2311.13384}, 2023.

\bibitem{curless1996volumetric}
B.~Curless and M.~Levoy.
\newblock A volumetric method for building complex models from range images.
\newblock In {\em Proceedings of the 23rd annual conference on Computer
  graphics and interactive techniques}, pages 303--312, 1996.

\bibitem{dai2017scannet}
A.~Dai, A.~X. Chang, M.~Savva, M.~Halber, T.~Funkhouser, and M.~Nie{\ss}ner.
\newblock Scannet: Richly-annotated 3d reconstructions of indoor scenes.
\newblock In {\em Proc. Computer Vision and Pattern Recognition (CVPR), IEEE},
  2017.

\bibitem{deitke2024objaverse}
M.~Deitke, R.~Liu, M.~Wallingford, H.~Ngo, O.~Michel, A.~Kusupati, A.~Fan,
  C.~Laforte, V.~Voleti, S.~Y. Gadre, et~al.
\newblock Objaverse-xl: A universe of 10m+ 3d objects.
\newblock {\em Advances in Neural Information Processing Systems}, 36, 2024.

\bibitem{deng2022gram}
Y.~Deng, J.~Yang, J.~Xiang, and X.~Tong.
\newblock Gram: Generative radiance manifolds for 3d-aware image generation.
\newblock In {\em IEEE Computer Vision and Pattern Recognition}, 2022.

\bibitem{devries2021unconstrained}
T.~Devries, M.~{\'A}. Bautista, N.~Srivastava, G.~W. Taylor, and J.~M.
  Susskind.
\newblock Unconstrained scene generation with locally conditioned radiance
  fields.
\newblock {\em 2021 IEEE/CVF International Conference on Computer Vision
  (ICCV)}, pages 14284--14293, 2021.

\bibitem{du2023cross}
Y.~Du, C.~Smith, A.~Tewari, and V.~Sitzmann.
\newblock Learning to render novel views from wide-baseline stereo pairs.
\newblock In {\em Proceedings of the IEEE/CVF Conference on Computer Vision and
  Pattern Recognition}, 2023.

\bibitem{eslami2018neural}
S.~A. Eslami, D.~J. Rezende, F.~Besse, F.~Viola, A.~S. Morcos, M.~Garnelo,
  A.~Ruderman, A.~A. Rusu, I.~Danihelka, K.~Gregor, et~al.
\newblock Neural scene representation and rendering.
\newblock {\em Science}, 360(6394):1204--1210, 2018.

\bibitem{esser2021taming}
P.~Esser, R.~Rombach, and B.~Ommer.
\newblock Taming transformers for high-resolution image synthesis.
\newblock In {\em Proceedings of the IEEE/CVF conference on computer vision and
  pattern recognition}, pages 12873--12883, 2021.

\bibitem{fridovich2023k}
S.~Fridovich-Keil, G.~Meanti, F.~R. Warburg, B.~Recht, and A.~Kanazawa.
\newblock K-planes: Explicit radiance fields in space, time, and appearance.
\newblock In {\em Proceedings of the IEEE/CVF Conference on Computer Vision and
  Pattern Recognition}, pages 12479--12488, 2023.

\bibitem{gao2020pile}
L.~Gao, S.~Biderman, S.~Black, L.~Golding, T.~Hoppe, C.~Foster, J.~Phang,
  H.~He, A.~Thite, N.~Nabeshima, S.~Presser, and C.~Leahy.
\newblock The {P}ile: An 800gb dataset of diverse text for language modeling.
\newblock {\em arXiv preprint arXiv:2101.00027}, 2020.

\bibitem{gao2024cat3d}
R.~Gao, A.~Holynski, P.~Henzler, A.~Brussee, R.~Martin-Brualla, P.~Srinivasan,
  J.~T. Barron, and B.~Poole.
\newblock Cat3d: Create anything in 3d with multi-view diffusion models.
\newblock {\em arXiv:2405.10314}, 2024.

\bibitem{goodfellow2014generative}
I.~Goodfellow, J.~Pouget-Abadie, M.~Mirza, B.~Xu, D.~Warde-Farley, S.~Ozair,
  A.~Courville, and Y.~Bengio.
\newblock Generative adversarial nets.
\newblock {\em Advances in neural information processing systems}, 27, 2014.

\bibitem{gortler1996lumigraph}
S.~J. Gortler, R.~Grzeszczuk, R.~Szeliski, and M.~F. Cohen.
\newblock The lumigraph.
\newblock In {\em Proceedings of the 23rd annual conference on Computer
  graphics and interactive techniques}, pages 43--54, 1996.

\bibitem{gu2023learning}
J.~Gu, Q.~Gao, S.~Zhai, B.~Chen, L.~Liu, and J.~Susskind.
\newblock Learning controllable 3d diffusion models from single-view images.
\newblock {\em ArXiv}, 2023.

\bibitem{gupta20233dgen}
A.~Gupta, W.~Xiong, Y.~Nie, I.~Jones, and B.~O{\u{g}}uz.
\newblock 3dgen: Triplane latent diffusion for textured mesh generation.
\newblock {\em arXiv preprint arXiv:2303.05371}, 2023.

\bibitem{hartley2003multiple}
R.~Hartley and A.~Zisserman.
\newblock {\em Multiple view geometry in computer vision}.
\newblock Cambridge university press, 2003.

\bibitem{henderson19ijcv}
P.~Henderson and V.~Ferrari.
\newblock Learning single-image {3D} reconstruction by generative modelling of
  shape, pose and shading.
\newblock {\em International Journal of Computer Vision (IJCV)}, 2019.

\bibitem{henderson21arxiv}
P.~Henderson, C.~H. Lampert, and B.~Bickel.
\newblock Unsupervised video prediction from a single frame by estimating 3d
  dynamic scene structure.
\newblock {\em CoRR}, abs/2106.09051, 2021.

\bibitem{henderson20cvpr}
P.~Henderson, V.~Tsiminaki, and C.~Lampert.
\newblock Leveraging {2D} data to learn textured {3D} mesh generation.
\newblock In {\em IEEE Conference on Computer Vision and Pattern Recognition
  (CVPR)}, 2020.

\bibitem{henzler2021unsupervised}
P.~Henzler, J.~Reizenstein, P.~Labatut, R.~Shapovalov, T.~Ritschel, A.~Vedaldi,
  and D.~Novotny.
\newblock Unsupervised learning of 3d object categories from videos in the
  wild.
\newblock In {\em Proceedings of the IEEE/CVF Conference on Computer Vision and
  Pattern Recognition}, pages 4700--4709, 2021.

\bibitem{ho2022cfg}
J.~Ho.
\newblock Classifier-free diffusion guidance.
\newblock {\em ArXiv}, abs/2207.12598, 2022.

\bibitem{ho2022imagen}
J.~Ho, W.~Chan, C.~Saharia, J.~Whang, R.~Gao, A.~Gritsenko, D.~P. Kingma,
  B.~Poole, M.~Norouzi, D.~J. Fleet, et~al.
\newblock Imagen video: High definition video generation with diffusion models.
\newblock {\em arXiv preprint arXiv:2210.02303}, 2022.

\bibitem{ho2020denoising}
J.~Ho, A.~Jain, and P.~Abbeel.
\newblock Denoising diffusion probabilistic models.
\newblock {\em Advances in Neural Information Processing Systems},
  33:6840--6851, 2020.

\bibitem{hoellein2024viewdiff}
L.~H{\"o}llein, A.~Bo\v{z}i\v{c}, N.~M{\"u}ller, D.~Novotny, H.-Y. Tseng,
  C.~Richardt, M.~Zollh{\"o}fer, and M.~Nie{\ss}ner.
\newblock Viewdiff: 3d-consistent image generation with text-to-image models.
\newblock In {\em Proceedings of the IEEE/CVF Conference on Computer Vision and
  Pattern Recognition}, 2024.

\bibitem{hu2024mvdfusion}
H.~Hu, Z.~Zhou, V.~Jampani, and S.~Tulsiani.
\newblock Mvd-fusion: Single-view 3d via depth-consistent multi-view
  generation.
\newblock In {\em CVPR}, 2024.

\bibitem{hui2022neural}
K.-H. Hui, R.~Li, J.~Hu, and C.-W. Fu.
\newblock Neural wavelet-domain diffusion for 3d shape generation.
\newblock In {\em SIGGRAPH Asia 2022 Conference Papers}, pages 1--9, 2022.

\bibitem{karnewar2023holodiffusion}
A.~Karnewar, A.~Vedaldi, D.~Novotny, and N.~Mitra.
\newblock Holodiffusion: Training a 3d diffusion model using 2d images.
\newblock {\em ArXiv}, 2023.

\bibitem{karras2019style}
T.~Karras, S.~Laine, and T.~Aila.
\newblock A style-based generator architecture for generative adversarial
  networks.
\newblock In {\em Proceedings of the IEEE/CVF Conference on Computer Vision and
  Pattern Recognition}, pages 4401--4410, 2019.

\bibitem{kerbl20233d}
B.~Kerbl, G.~Kopanas, T.~Leimk{\"u}hler, and G.~Drettakis.
\newblock 3d gaussian splatting for real-time radiance field rendering.
\newblock {\em ACM Transactions on Graphics}, 42(4):1--14, 2023.

\bibitem{kim2023neuralfieldldm}
S.~W. Kim, B.~Brown, K.~Yin, K.~Kreis, K.~Schwarz, D.~Li, R.~Rombach,
  A.~Torralba, and S.~Fidler.
\newblock Neuralfield-ldm: Scene generation with hierarchical latent diffusion
  models.
\newblock In {\em IEEE Conference on Computer Vision and Pattern Recognition
  ({CVPR})}, 2023.

\bibitem{kingma2013auto}
D.~P. Kingma and M.~Welling.
\newblock Auto-encoding variational bayes.
\newblock In Y.~Bengio and Y.~LeCun, editors, {\em 2nd International Conference
  on Learning Representations, {ICLR} 2014, Banff, AB, Canada, April 14-16,
  2014, Conference Track Proceedings}, 2014.

\bibitem{kong2024eschernet}
X.~Kong, S.~Liu, X.~Lyu, M.~Taher, X.~Qi, and A.~J. Davison.
\newblock Eschernet: A generative model for scalable view synthesis.
\newblock {\em arXiv preprint arXiv:2402.03908}, 2024.

\bibitem{kosiorek2021nerf}
A.~R. Kosiorek, H.~Strathmann, D.~Zoran, P.~Moreno, R.~Schneider,
  S.~Mokr{\'{a}}, and D.~J. Rezende.
\newblock Nerf-vae: {A} geometry aware 3d scene generative model.
\newblock In M.~Meila and T.~Zhang, editors, {\em Proceedings of the 38th
  International Conference on Machine Learning, {ICML} 2021, 18-24 July 2021,
  Virtual Event}, volume 139 of {\em Proceedings of Machine Learning Research},
  pages 5742--5752. {PMLR}, 2021.

\bibitem{kulhanek2022viewformer}
J.~Kulh{\'a}nek, E.~Derner, T.~Sattler, and R.~Babu{\v{s}}ka.
\newblock Viewformer: Nerf-free neural rendering from few images using
  transformers.
\newblock In {\em European Conference on Computer Vision (ECCV)}, 2022.

\bibitem{li2022diffusionsdf}
M.~Li, Y.~Duan, J.~Zhou, and J.~Lu.
\newblock Diffusion-sdf: Text-to-shape via voxelized diffusion.
\newblock In {\em Proceedings of the IEEE Conference on Computer Vision and
  Pattern Recognition (CVPR)}, 2023.

\bibitem{li2023neuralangelo}
Z.~Li, T.~M\"uller, A.~Evans, R.~H. Taylor, M.~Unberath, M.-Y. Liu, and C.-H.
  Lin.
\newblock Neuralangelo: High-fidelity neural surface reconstruction.
\newblock In {\em IEEE Conference on Computer Vision and Pattern Recognition
  ({CVPR})}, 2023.

\bibitem{lin2023magic3d}
C.-H. Lin, J.~Gao, L.~Tang, T.~Takikawa, X.~Zeng, X.~Huang, K.~Kreis,
  S.~Fidler, M.-Y. Liu, and T.-Y. Lin.
\newblock Magic3d: High-resolution text-to-3d content creation.
\newblock In {\em IEEE Conference on Computer Vision and Pattern Recognition
  ({CVPR})}, 2023.

\bibitem{liu2024one}
M.~Liu, C.~Xu, H.~Jin, L.~Chen, M.~Varma~T, Z.~Xu, and H.~Su.
\newblock One-2-3-45: Any single image to 3d mesh in 45 seconds without
  per-shape optimization.
\newblock {\em Advances in Neural Information Processing Systems}, 36, 2024.

\bibitem{liu2023zero1to3}
R.~Liu, R.~Wu, B.~V. Hoorick, P.~Tokmakov, S.~Zakharov, and C.~Vondrick.
\newblock Zero-1-to-3: Zero-shot one image to 3d object.
\newblock {\em ArXiv}, 2023.

\bibitem{liu2023syncdreamer}
Y.~Liu, C.~Lin, Z.~Zeng, X.~Long, L.~Liu, T.~Komura, and W.~Wang.
\newblock Syncdreamer: Generating multiview-consistent images from a
  single-view image.
\newblock In {\em The Twelfth International Conference on Learning
  Representations}, 2024.

\bibitem{liu2022neural}
Y.~Liu, S.~Peng, L.~Liu, Q.~Wang, P.~Wang, C.~Theobalt, X.~Zhou, and W.~Wang.
\newblock Neural rays for occlusion-aware image-based rendering.
\newblock In {\em Proceedings of the IEEE/CVF Conference on Computer Vision and
  Pattern Recognition}, pages 7824--7833, 2022.

\bibitem{adamw}
I.~Loshchilov and F.~Hutter.
\newblock Decoupled weight decay regularization.
\newblock In {\em International Conference on Learning Representations (ICLR)},
  2019.

\bibitem{pointclouddiffusion}
S.~Luo and W.~Hu.
\newblock Diffusion probabilistic models for 3d point cloud generation.
\newblock {\em 2021 IEEE/CVF Conference on Computer Vision and Pattern
  Recognition (CVPR)}, Jun 2021.

\bibitem{melaskyriazi2023realfusion}
L.~Melas-Kyriazi, C.~Rupprecht, I.~Laina, and A.~Vedaldi.
\newblock Realfusion: 360° reconstruction of any object from a single image.
\newblock In {\em Arxiv}, 2023.

\bibitem{mildenhall2020nerf}
B.~Mildenhall, P.~P. Srinivasan, M.~Tancik, J.~T. Barron, R.~Ramamoorthi, and
  R.~Ng.
\newblock Nerf: Representing scenes as neural radiance fields for view
  synthesis.
\newblock In {\em ECCV}, 2020.

\bibitem{mueller2022diffrf}
N.~M{\"u}ller, Y.~Siddiqui, L.~Porzi, S.~R. Bulo, P.~Kontschieder, and
  M.~Nie{\ss}ner.
\newblock Diffrf: Rendering-guided 3d radiance field diffusion.
\newblock In {\em Proceedings of the IEEE/CVF Conference on Computer Vision and
  Pattern Recognition}, pages 4328--4338, 2023.

\bibitem{muller2022instant}
T.~M\"uller, A.~Evans, C.~Schied, and A.~Keller.
\newblock Instant neural graphics primitives with a multiresolution hash
  encoding.
\newblock {\em ACM Trans. Graph.}, 41(4):102:1--102:15, July 2022.

\bibitem{hologan}
T.~Nguyen-Phuoc, C.~Li, L.~Theis, C.~Richardt, and Y.-L. Yang.
\newblock Hologan: Unsupervised learning of 3d representations from natural
  images.
\newblock In {\em Proceedings of the IEEE/CVF International Conference on
  Computer Vision}, pages 7588--7597, 2019.

\bibitem{nguyen2020blockgan}
T.~Nguyen-Phuoc, C.~Richardt, L.~Mai, Y.-L. Yang, and N.~Mitra.
\newblock Blockgan: Learning 3d object-aware scene representations from
  unlabelled images.
\newblock In {\em Advances in Neural Information Processing Systems 33}, Nov
  2020.

\bibitem{regnerf2022}
M.~Niemeyer, J.~T. Barron, B.~Mildenhall, M.~S.~M. Sajjadi, A.~Geiger, and
  N.~Radwan.
\newblock Regnerf: Regularizing neural radiance fields for view synthesis from
  sparse inputs.
\newblock {\em 2022 IEEE/CVF Conference on Computer Vision and Pattern
  Recognition (CVPR)}, Jun 2022.

\bibitem{park2019deepsdf}
J.~J. Park, P.~Florence, J.~Straub, R.~Newcombe, and S.~Lovegrove.
\newblock Deepsdf: Learning continuous signed distance functions for shape
  representation.
\newblock In {\em Proceedings of the IEEE/CVF conference on computer vision and
  pattern recognition}, pages 165--174, 2019.

\bibitem{peng2020convolutional}
S.~Peng, M.~Niemeyer, L.~Mescheder, M.~Pollefeys, and A.~Geiger.
\newblock Convolutional occupancy networks.
\newblock In {\em Computer Vision--ECCV 2020: 16th European Conference,
  Glasgow, UK, August 23--28, 2020, Proceedings, Part III 16}, pages 523--540.
  Springer, 2020.

\bibitem{poole2022dreamfusion}
B.~Poole, A.~Jain, J.~T. Barron, and B.~Mildenhall.
\newblock Dreamfusion: Text-to-3d using 2d diffusion.
\newblock {\em ArXiv}, 2022.

\bibitem{radford2018improving}
A.~Radford and K.~Narasimhan.
\newblock Improving language understanding by generative pre-training.
\newblock OpenAI, 2018.

\bibitem{reizenstein21co3d}
J.~Reizenstein, R.~Shapovalov, P.~Henzler, L.~Sbordone, P.~Labatut, and
  D.~Novotny.
\newblock Common objects in 3d: Large-scale learning and evaluation of
  real-life 3d category reconstruction.
\newblock In {\em International Conference on Computer Vision}, 2021.

\bibitem{rezende2014stochastic}
D.~J. Rezende, S.~Mohamed, and D.~Wierstra.
\newblock Stochastic backpropagation and approximate inference in deep
  generative models.
\newblock In {\em International conference on machine learning}, pages
  1278--1286. PMLR, 2014.

\bibitem{rockwell2021pixelsynth}
C.~Rockwell, D.~F. Fouhey, and J.~Johnson.
\newblock Pixelsynth: Generating a 3d-consistent experience from a single
  image.
\newblock In {\em ICCV}, 2021.

\bibitem{roessle2023ganerf}
B.~Roessle, N.~M{\"u}ller, L.~Porzi, S.~R. Bul{\`o}, P.~Kontschieder, and
  M.~Nie{\ss}ner.
\newblock Ganerf: Leveraging discriminators to optimize neural radiance fields.
\newblock {\em ACM Trans. Graph.}, 42(6), nov 2023.

\bibitem{Rombach_2022_CVPR}
R.~Rombach, A.~Blattmann, D.~Lorenz, P.~Esser, and B.~Ommer.
\newblock High-resolution image synthesis with latent diffusion models.
\newblock In {\em Proceedings of the IEEE/CVF Conference on Computer Vision and
  Pattern Recognition (CVPR)}, pages 10684--10695, June 2022.

\bibitem{rombach2021geometryfree}
R.~Rombach, P.~Esser, and B.~Ommer.
\newblock Geometry-free view synthesis: Transformers and no 3d priors.
\newblock {\em ArXiv}, 2021.

\bibitem{RonnebergerFB15}
O.~Ronneberger, P.~Fischer, and T.~Brox.
\newblock U-net: Convolutional networks for biomedical image segmentation.
\newblock In {\em MICCAI}, 2015.

\bibitem{salimans2022vpred}
T.~Salimans and J.~Ho.
\newblock Progressive distillation for fast sampling of diffusion models.
\newblock In {\em International Conference on Learning Representations (ICLR)},
  2022.

\bibitem{schoenberger2016sfm}
J.~L. Sch\"{o}nberger and J.-M. Frahm.
\newblock Structure-from-motion revisited.
\newblock In {\em Conference on Computer Vision and Pattern Recognition
  (CVPR)}, 2016.

\bibitem{schuhmann2022laion}
C.~Schuhmann, R.~Beaumont, R.~Vencu, C.~Gordon, R.~Wightman, M.~Cherti,
  T.~Coombes, A.~Katta, C.~Mullis, M.~Wortsman, et~al.
\newblock Laion-5b: An open large-scale dataset for training next generation
  image-text models.
\newblock {\em Advances in Neural Information Processing Systems},
  35:25278--25294, 2022.

\bibitem{schwarz2020graf}
K.~Schwarz, Y.~Liao, M.~Niemeyer, and A.~Geiger.
\newblock {GRAF:} generative radiance fields for 3d-aware image synthesis.
\newblock In H.~Larochelle, M.~Ranzato, R.~Hadsell, M.~Balcan, and H.~Lin,
  editors, {\em Advances in Neural Information Processing Systems 33: Annual
  Conference on Neural Information Processing Systems 2020, NeurIPS 2020,
  December 6-12, 2020, virtual}, 2020.

\bibitem{Schwarz2024ICLR}
K.~Schwarz, S.~Wook~Kim, J.~Gao, S.~Fidler, A.~Geiger, and K.~Kreis.
\newblock Wildfusion: Learning 3d-aware latent diffusion models in view space.
\newblock In {\em International Conference on Learning Representations (ICLR)},
  2024.

\bibitem{seitz2006comparison}
S.~M. Seitz, B.~Curless, J.~Diebel, D.~Scharstein, and R.~Szeliski.
\newblock A comparison and evaluation of multi-view stereo reconstruction
  algorithms.
\newblock In {\em 2006 IEEE computer society conference on computer vision and
  pattern recognition (CVPR'06)}, volume~1, pages 519--528. IEEE, 2006.

\bibitem{seitz1999photorealistic}
S.~M. Seitz and C.~R. Dyer.
\newblock Photorealistic scene reconstruction by voxel coloring.
\newblock {\em International journal of computer vision}, 35:151--173, 1999.

\bibitem{shen2024gamba}
Q.~Shen, X.~Yi, Z.~Wu, P.~Zhou, H.~Zhang, S.~Yan, and X.~Wang.
\newblock Gamba: Marry gaussian splatting with mamba for single view 3d
  reconstruction.
\newblock {\em ArXiv}, 2024.

\bibitem{shriram2024realmdreamer}
J.~Shriram, A.~Trevithick, L.~Liu, and R.~Ramamoorthi.
\newblock {RealmDreamer}: Text-driven 3d scene generation with inpainting and
  depth diffusion.
\newblock {\em ArXiv}, 2024.

\bibitem{shue20223d}
J.~R. Shue, E.~R. Chan, R.~Po, Z.~Ankner, J.~Wu, and G.~Wetzstein.
\newblock 3d neural field generation using triplane diffusion.
\newblock {\em arXiv preprint arXiv:2211.16677}, 2022.

\bibitem{sitzmann2019deepvoxels}
V.~Sitzmann, J.~Thies, F.~Heide, M.~Nie{\ss}ner, G.~Wetzstein, and
  M.~Zollh{\"o}fer.
\newblock Deepvoxels: Learning persistent 3d feature embeddings.
\newblock In {\em Proc. Computer Vision and Pattern Recognition (CVPR), IEEE},
  2019.

\bibitem{skorokhodov2022epigraf}
I.~Skorokhodov, S.~Tulyakov, Y.~Wang, and P.~Wonka.
\newblock Epigraf: Rethinking training of 3d gans.
\newblock In S.~Koyejo, S.~Mohamed, A.~Agarwal, D.~Belgrave, K.~Cho, and A.~Oh,
  editors, {\em Advances in Neural Information Processing Systems}, volume~35,
  pages 24487--24501. Curran Associates, Inc., 2022.

\bibitem{snavely2006photo}
N.~Snavely, S.~M. Seitz, and R.~Szeliski.
\newblock Photo tourism: exploring photo collections in 3d.
\newblock In {\em ACM siggraph 2006 papers}, pages 835--846. 2006.

\bibitem{DBLP:journals/corr/Sohl-DicksteinW15}
J.~Sohl{-}Dickstein, E.~A. Weiss, N.~Maheswaranathan, and S.~Ganguli.
\newblock Deep unsupervised learning using nonequilibrium thermodynamics.
\newblock {\em CoRR}, abs/1503.03585, 2015.

\bibitem{song2020denoising}
J.~Song, C.~Meng, and S.~Ermon.
\newblock Denoising diffusion implicit models.
\newblock In {\em International Conference on Learning Representations}, 2021.

\bibitem{szymanowicz2023viewset_diffusion}
S.~Szymanowicz, C.~Rupprecht, and A.~Vedaldi.
\newblock Viewset diffusion: (0-)image-conditioned {3D} generative models from
  {2D} data.
\newblock In {\em ICCV}, 2023.

\bibitem{szymanowicz2024splatter_image}
S.~Szymanowicz, C.~Rupprecht, and A.~Vedaldi.
\newblock Splatter image: Ultra-fast single-view 3d reconstruction.
\newblock {\em Conference on Computer Vision and Pattern Recognition (CVPR)},
  2024.

\bibitem{tang2024lgm}
J.~Tang, Z.~Chen, X.~Chen, T.~Wang, G.~Zeng, and Z.~Liu.
\newblock Lgm: Large multi-view gaussian model for high-resolution 3d content
  creation.
\newblock {\em arXiv preprint arXiv:2402.05054}, 2024.

\bibitem{tang2023dreamgaussian}
J.~Tang, J.~Ren, H.~Zhou, Z.~Liu, and G.~Zeng.
\newblock Dreamgaussian: Generative gaussian splatting for efficient 3d content
  creation.
\newblock In {\em The Twelfth International Conference on Learning
  Representations}, 2024.

\bibitem{tang2024mvdiffusionpp}
S.~Tang, J.~Chen, D.~Wang, C.~Tang, F.~Zhang, Y.~Fan, V.~Chandra, Y.~Furukawa,
  and R.~Ranjan.
\newblock Mvdiffusion++: A dense high-resolution multi-view diffusion model for
  single or sparse-view 3d object reconstruction.
\newblock {\em arXiv preprint arXiv:2402.12712}, 2024.

\bibitem{Tang2023mvdiffusion}
S.~Tang, F.~Zhang, J.~Chen, P.~Wang, and Y.~Furukawa.
\newblock {MVD}iffusion: Enabling holistic multi-view image generation with
  correspondence-aware diffusion.
\newblock In {\em Thirty-seventh Conference on Neural Information Processing
  Systems}, 2023.

\bibitem{tewari2023forwarddiffusion}
A.~Tewari, T.~Yin, G.~Cazenavette, S.~Rezchikov, J.~B. Tenenbaum, F.~Durand,
  W.~T. Freeman, and V.~Sitzmann.
\newblock Diffusion with forward models: Solving stochastic inverse problems
  without direct supervision.
\newblock In {\em Thirty-seventh Conference on Neural Information Processing
  Systems}, 2023.

\bibitem{tseng2023consistent}
H.-Y. Tseng, Q.~Li, C.~Kim, S.~Alsisan, J.-B. Huang, and J.~Kopf.
\newblock Consistent view synthesis with pose-guided diffusion models.
\newblock {\em ArXiv}, 2023.

\bibitem{vahdat2022lion}
A.~Vahdat, F.~Williams, Z.~Gojcic, O.~Litany, S.~Fidler, K.~Kreis, et~al.
\newblock Lion: Latent point diffusion models for 3d shape generation.
\newblock {\em Advances in Neural Information Processing Systems},
  35:10021--10039, 2022.

\bibitem{oord2016wavenet}
A.~van~den Oord, S.~Dieleman, H.~Zen, K.~Simonyan, O.~Vinyals, A.~Graves,
  N.~Kalchbrenner, A.~Senior, and K.~Kavukcuoglu.
\newblock Wavenet: A generative model for raw audio.
\newblock In {\em Arxiv}, 2016.

\bibitem{van2016pixel}
A.~Van Den~Oord, N.~Kalchbrenner, and K.~Kavukcuoglu.
\newblock Pixel recurrent neural networks.
\newblock In {\em International conference on machine learning}, pages
  1747--1756. PMLR, 2016.

\bibitem{vaswani17attention}
A.~Vaswani, N.~Shazeer, N.~Parmar, J.~Uszkoreit, L.~Jones, A.~N. Gomez, L.~u.
  Kaiser, and I.~Polosukhin.
\newblock Attention is all you need.
\newblock In I.~Guyon, U.~V. Luxburg, S.~Bengio, H.~Wallach, R.~Fergus,
  S.~Vishwanathan, and R.~Garnett, editors, {\em Advances in Neural Information
  Processing Systems}, volume~30. Curran Associates, Inc., 2017.

\bibitem{wang2022score}
H.~Wang, X.~Du, J.~Li, R.~A. Yeh, and G.~Shakhnarovich.
\newblock Score jacobian chaining: Lifting pretrained 2d diffusion models for
  3d generation.
\newblock {\em ArXiv}, 2022.

\bibitem{wang2023score}
H.~Wang, X.~Du, J.~Li, R.~A. Yeh, and G.~Shakhnarovich.
\newblock Score jacobian chaining: Lifting pretrained 2d diffusion models for
  3d generation.
\newblock In {\em Proceedings of the IEEE/CVF Conference on Computer Vision and
  Pattern Recognition}, pages 12619--12629, 2023.

\bibitem{wang2021ibrnet}
Q.~Wang, Z.~Wang, K.~Genova, P.~Srinivasan, H.~Zhou, J.~T. Barron,
  R.~Martin-Brualla, N.~Snavely, and T.~Funkhouser.
\newblock Ibrnet: Learning multi-view image-based rendering.
\newblock In {\em CVPR}, 2021.

\bibitem{wang2022rodin}
T.~Wang, B.~Zhang, T.~Zhang, S.~Gu, J.~Bao, T.~Baltrusaitis, J.~Shen, D.~Chen,
  F.~Wen, Q.~Chen, and B.~Guo.
\newblock Rodin: A generative model for sculpting 3d digital avatars using
  diffusion.
\newblock {\em ArXiv}, 2022.

\bibitem{wang2024prolificdreamer}
Z.~Wang, C.~Lu, Y.~Wang, F.~Bao, C.~Li, H.~Su, and J.~Zhu.
\newblock Prolificdreamer: High-fidelity and diverse text-to-3d generation with
  variational score distillation.
\newblock {\em Advances in Neural Information Processing Systems}, 36, 2024.

\bibitem{watson2022novel}
D.~Watson, W.~Chan, R.~M. Brualla, J.~Ho, A.~Tagliasacchi, and M.~Norouzi.
\newblock Novel view synthesis with diffusion models.
\newblock In {\em The Eleventh International Conference on Learning
  Representations}, 2023.

\bibitem{wewer24latentsplat}
C.~Wewer, K.~Raj, E.~Ilg, B.~Schiele, and J.~E. Lenssen.
\newblock latentsplat: Autoencoding variational gaussians for fast
  generalizable 3d reconstruction.
\newblock In {\em arXiv}, 2024.

\bibitem{wiles2020synsin}
O.~Wiles, G.~Gkioxari, R.~Szeliski, and J.~Johnson.
\newblock Synsin: End-to-end view synthesis from a single image.
\newblock In {\em Proceedings of the IEEE/CVF Conference on Computer Vision and
  Pattern Recognition}, pages 7467--7477, 2020.

\bibitem{wu2023multiview}
C.-Y. Wu, J.~Johnson, J.~Malik, C.~Feichtenhofer, and G.~Gkioxari.
\newblock Multiview compressive coding for 3d reconstruction.
\newblock In {\em Proceedings of the IEEE/CVF Conference on Computer Vision and
  Pattern Recognition}, pages 9065--9075, 2023.

\bibitem{wu2023reconfusion}
R.~Wu, B.~Mildenhall, P.~Henzler, K.~Park, R.~Gao, D.~Watson, P.~P. Srinivasan,
  D.~Verbin, J.~T. Barron, B.~Poole, and A.~Holynski.
\newblock Reconfusion: 3d reconstruction with diffusion priors.
\newblock {\em ArXiv}, 2023.

\bibitem{wu2023omniobject3d}
T.~Wu, J.~Zhang, X.~Fu, Y.~Wang, L.~P. Jiawei~Ren, W.~Wu, L.~Yang, J.~Wang,
  C.~Qian, D.~Lin, and Z.~Liu.
\newblock Omniobject3d: Large-vocabulary 3d object dataset for realistic
  perception, reconstruction and generation.
\newblock In {\em IEEE/CVF Conference on Computer Vision and Pattern
  Recognition (CVPR)}, 2023.

\bibitem{wynn-2023-diffusionerf}
J.~Wynn and D.~Turmukhambetov.
\newblock {DiffusioNeRF: Regularizing Neural Radiance Fields with Denoising
  Diffusion Models}.
\newblock In {\em CVPR}, 2023.

\bibitem{xie22neuralfields}
Y.~Xie, T.~Takikawa, S.~Saito, O.~Litany, S.~Yan, N.~Khan, F.~Tombari,
  J.~Tompkin, V.~Sitzmann, and S.~Sridhar.
\newblock Neural fields in visual computing and beyond.
\newblock {\em Computer Graphics Forum}, 2022.

\bibitem{xu2022point}
Q.~Xu, Z.~Xu, J.~Philip, S.~Bi, Z.~Shu, K.~Sunkavalli, and U.~Neumann.
\newblock Point-nerf: Point-based neural radiance fields.
\newblock In {\em Proceedings of the IEEE/CVF Conference on Computer Vision and
  Pattern Recognition}, pages 5438--5448, 2022.

\bibitem{xu2023dmv3d}
Y.~Xu, H.~Tan, F.~Luan, S.~Bi, P.~Wang, J.~Li, Z.~Shi, K.~Sunkavalli,
  G.~Wetzstein, Z.~Xu, and K.~Zhang.
\newblock {DMV}3d: Denoising multi-view diffusion using 3d large reconstruction
  model.
\newblock In {\em The Twelfth International Conference on Learning
  Representations}, 2024.

\bibitem{yi2023gaussiandreamer}
T.~Yi, J.~Fang, J.~Wang, G.~Wu, L.~Xie, X.~Zhang, W.~Liu, Q.~Tian, and X.~Wang.
\newblock Gaussiandreamer: Fast generation from text to 3d gaussians by
  bridging 2d and 3d diffusion models.
\newblock In {\em CVPR}, 2024.

\bibitem{xu2024grm}
X.~Yinghao, S.~Zifan, Y.~Wang, C.~Hansheng, Y.~Ceyuan, P.~Sida, S.~Yujun, and
  W.~Gordon.
\newblock Grm: Large gaussian reconstruction model for efficient 3d
  reconstruction and generation.
\newblock {\em ArXiv}, 2024.

\bibitem{yoo2023dreamsparse}
P.~Yoo, J.~Guo, Y.~Matsuo, and S.~S. Gu.
\newblock Dreamsparse: Escaping from plato's cave with 2d frozen diffusion
  model given sparse views.
\newblock {\em CoRR}, 2023.

\bibitem{yu2021pixelnerf}
A.~Yu, V.~Ye, M.~Tancik, and A.~Kanazawa.
\newblock pixelnerf: Neural radiance fields from one or few images.
\newblock In {\em Proceedings of the IEEE/CVF Conference on Computer Vision and
  Pattern Recognition}, pages 4578--4587, 2021.

\bibitem{yu2023longterm}
J.~J. Yu, F.~Forghani, K.~G. Derpanis, and M.~A. Brubaker.
\newblock Long-term photometric consistent novel view synthesis with diffusion
  models.
\newblock In {\em {Proceedings of the International Conference on Computer
  Vision ({ICCV})}}, 2023.

\bibitem{yu2023mvimgnet}
X.~Yu, M.~Xu, Y.~Zhang, H.~Liu, C.~Ye, Y.~Wu, Z.~Yan, T.~Liang, G.~Chen,
  S.~Cui, and X.~Han.
\newblock Mvimgnet: A large-scale dataset of multi-view images.
\newblock In {\em CVPR}, 2023.

\bibitem{zhang2024gaussiancube}
B.~Zhang, Y.~Cheng, J.~Yang, C.~Wang, F.~Zhao, Y.~Tang, D.~Chen, and B.~Guo.
\newblock Gaussiancube: Structuring gaussian splatting using optimal transport
  for 3d generative modeling.
\newblock {\em arXiv preprint arXiv:2403.19655}, 2024.

\bibitem{gslrm2024}
K.~Zhang, S.~Bi, H.~Tan, Y.~Xiangli, N.~Zhao, K.~Sunkavalli, and Z.~Xu.
\newblock Gs-lrm: Large reconstruction model for 3d gaussian splatting.
\newblock {\em ArXiv}, 2024.

\bibitem{zhang2018unreasonable}
R.~Zhang, P.~Isola, A.~A. Efros, E.~Shechtman, and O.~Wang.
\newblock The unreasonable effectiveness of deep features as a perceptual
  metric.
\newblock In {\em Proceedings of the IEEE conference on computer vision and
  pattern recognition}, pages 586--595, 2018.

\bibitem{gmpi2022}
X.~Zhao, F.~Ma, D.~Güera, Z.~Ren, A.~G. Schwing, and A.~Colburn.
\newblock Generative multiplane images: Making a 2d gan 3d-aware.
\newblock In {\em Proc. ECCV}, 2022.

\bibitem{zheng2024gpsgaussian}
S.~Zheng, B.~Zhou, R.~Shao, B.~Liu, S.~Zhang, L.~Nie, and Y.~Liu.
\newblock Gps-gaussian: Generalizable pixel-wise 3d gaussian splatting for
  real-time human novel view synthesis.
\newblock In {\em Proceedings of the IEEE/CVF Conference on Computer Vision and
  Pattern Recognition (CVPR)}, 2024.

\bibitem{zhou20213d}
L.~Zhou, Y.~Du, and J.~Wu.
\newblock 3d shape generation and completion through point-voxel diffusion.
\newblock In {\em Proceedings of the IEEE/CVF International Conference on
  Computer Vision}, pages 5826--5835, 2021.

\bibitem{zhou2018stereomag}
T.~Zhou, R.~Tucker, J.~Flynn, G.~Fyffe, and N.~Snavely.
\newblock Stereo magnification: Learning view synthesis using multiplane
  images.
\newblock {\em ACM Trans. Graph. (Proc. SIGGRAPH)}, 37, 2018.

\bibitem{zhou2023sparsefusion}
Z.~Zhou and S.~Tulsiani.
\newblock Sparsefusion: Distilling view-conditioned diffusion for 3d
  reconstruction.
\newblock In {\em CVPR}, 2023.

\bibitem{zhu2023hifa}
J.~Zhu and P.~Zhuang.
\newblock Hifa: High-fidelity text-to-3d with advanced diffusion guidance.
\newblock {\em arXiv preprint arXiv:2305.18766}, 2023.

\bibitem{zou2023sparse3d}
Z.-X. Zou, W.~Cheng, Y.-P. Cao, S.-S. Huang, Y.~Shan, and S.-H. Zhang.
\newblock Sparse3d: Distilling multiview-consistent diffusion for object
  reconstruction from sparse views.
\newblock {\em arXiv preprint arXiv:2308.14078}, 2023.

\bibitem{zou2023triplane}
Z.-X. Zou, Z.~Yu, Y.-C. Guo, Y.~Li, D.~Liang, Y.-P. Cao, and S.-H. Zhang.
\newblock Triplane meets gaussian splatting: Fast and generalizable single-view
  3d reconstruction with transformers.
\newblock {\em ArXiv}, 2023.

\end{thebibliography}
}
\clearpage

\appendix

\section{Implementation Details}

\paragraph{Optimisation.}
We train using AdamW~\cite{adamw} with a cosine learning rate schedule having maximum value $6\times10^{-5}$ and 500 steps for warm-up.
For regularisation, we use weight decay with strength $4\times10^{-2}$, and dropout with probability 0.28 after each U-Net block.
Total batch size is 24 for the autoencoder and 64 for the denoiser.
For the denoiser, we use a linear noise schedule over 1000 steps, with $v$-prediction as the objective~\cite{salimans2022vpred}.
These hyperparameters were chosen using automated random sweeps, choosing the best-performing based on validation-set PSNR for single-image reconstruction.
During autoencoder training the KL weight $\beta$ is set to 0.1, and we give the L2 and LPIPS losses equal weight.
We train with class conditioning (or unconditional for RealEstate10K) for 40\% of batches, image conditioning (for single-image reconstruction) for 40\% of batches, and no conditioning (to enable CFG) for the remaining 20\%.

\paragraph{View selection.}
During autoencoder training, we select groups of six frames from the videos as input to our model; these are drawn with stratified sampling so they are roughly evenly spaced through the entire original video clip, minimising visual ambiguity.
The autoencoder is trained to reconstruct six immediately-adjacent frames (each randomly chosen as preceding or following an input frame); this discourages pathological solutions with trivial 3D geometry that can arise when reconstructing the exact input frames.
During denoiser training, we again sample sets of input views following the same strategy. No disjoint target views are required for this stage since pathological solutions do not arise in the latent space.
When training with image conditioning, we randomly select one view to use as the conditioning image, and pass this separately through the encoder $E$.
For testing single-image reconstruction, we sample 12 views per scene. The middle view was used as the input for MVImgNet and the first view for RealEstate10K; reconstruction metrics are calculated on the remaining 11 views.
For testing sparse-view reconstruction, we sample 12 views per scene, using alternate views for input and evaluation.
For testing unconditional/class-conditional generation, we use randomly chosen sets of six camera poses from the validation set.

\paragraph{Splatter Image.}
We utilized the publicly available implementation from the original authors, training it on the same splits of MVImgNet and RealEstate10K as our model. During training, we processed mini-batches of scenes, each consisting of six posed views selected as for our model. One view was randomly chosen as the input, while the remaining views served as ground truth to optimize the model by comparing the generated renderings with the true images. For both datasets, we enable prediction of 3D splat offsets.
For testing, the middle view was used as the input for MVImgNet, and the first view for RealEstate10K.

\section{Additional Results}

\subsection{Denoised vs heldout views}
Here we measure how the quality of generated scenes varies between `denoised' and `held-out' views. The denoised views are those on which we support both the splats and the latent representation $z_v$; the held-out views are evenly spaced between these (i.e.~as far from them as possible).
It is natural to consider whether our image-centric approach encourages higher-quality images from the denoised viewpoints.
However, we find (see Table~\ref{tab:diff-vs-heldout-results}) that there is only minimal difference---FID is slightly better on diffused views for both datasets, while held-out views actually show fractionally better reconstruction metrics.

\subsection{Additional Qualitative results}

\begin{figure}[htbp]
\centering
    \begin{subfigure}{\textwidth}
        \resizebox{\textwidth}{!}{%
            \begin{tikzpicture}
              \node[inner sep=0] (image1) {\includegraphics[height=5cm]{figures/gen/mvimgnet/0000603a_00.png}};
              \node[rotate=90, anchor=south, font=\LARGE] at (image1.west) {\texttt{\textcolor{GHOST}{p}car\textcolor{GHOST}{p}}};  
            \end{tikzpicture}
            \begin{tikzpicture}
              \node[inner sep=0] (image2) {\includegraphics[height=5cm]{figures/gen/mvimgnet/2700a898_03.png}};
              \node[rotate=90, anchor=south, font=\LARGE] at (image2.west) {\texttt{apple}};  
            \end{tikzpicture}%
        }
        \resizebox{\textwidth}{!}{%
            \begin{tikzpicture}
              \node[inner sep=0] (image1) {\includegraphics[height=5cm]{figures/gen/mvimgnet/1d00f58d_02.png}};
              \node[rotate=90, anchor=south, font=\LARGE] at (image1.west) {\texttt{\textcolor{GHOST}{p}sunflower\textcolor{GHOST}{p}}};  
            \end{tikzpicture}
            \begin{tikzpicture}
              \node[inner sep=0] (image2) {\includegraphics[height=5cm]{figures/gen/mvimgnet/18009ad1_02.png}};
              \node[rotate=90, anchor=south, font=\LARGE] at (image2.west) {\texttt{pressure cooker}};  
            \end{tikzpicture}%
        }
        \resizebox{\textwidth}{!}{%
            \begin{tikzpicture}
              \node[inner sep=0] (image1) {\includegraphics[height=5cm]{figures/gen/mvimgnet/17007c04_02.png}};
              \node[rotate=90, anchor=south, font=\LARGE] at (image1.west) {\texttt{pineapple}};  
            \end{tikzpicture}
            \begin{tikzpicture}
              \node[inner sep=0] (image2) {\includegraphics[height=5cm]{figures/gen/mvimgnet/2500926c_00.png}};
              \node[rotate=90, anchor=south, font=\LARGE] at (image2.west) {\texttt{pants}};  
            \end{tikzpicture}%
        }
        \resizebox{\textwidth}{!}{%
            \begin{tikzpicture}
              \node[inner sep=0] (image1) {\includegraphics[height=5cm]{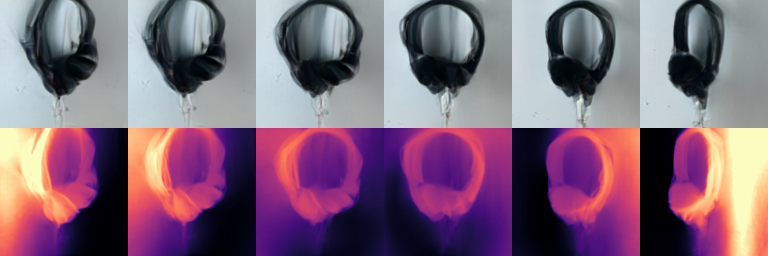}};
              \node[rotate=90, anchor=south, font=\LARGE] at (image1.west) {\texttt{earphones}};  
            \end{tikzpicture}
            \begin{tikzpicture}
              \node[inner sep=0] (image2) {\includegraphics[height=5cm]{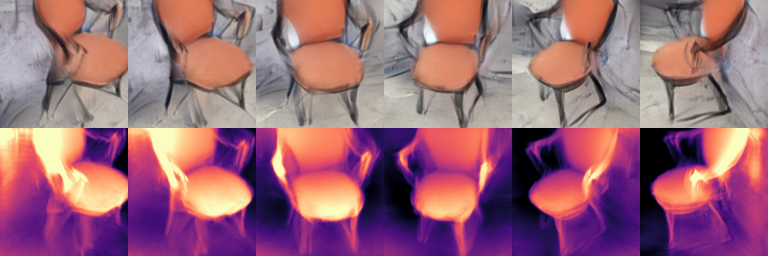}};
              \node[rotate=90, anchor=south, font=\LARGE] at (image2.west) {\texttt{\textcolor{GHOST}{p}chair\textcolor{GHOST}{p}}};  
            \end{tikzpicture}%
        }
        \resizebox{\textwidth}{!}{%
            \begin{tikzpicture}
              \node[inner sep=0] (image1) {\includegraphics[height=5cm]{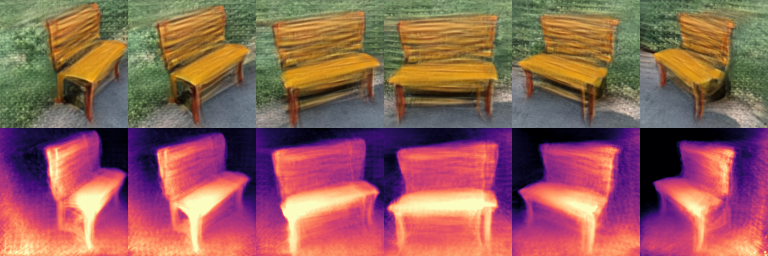}};
              \node[rotate=90, anchor=south, font=\LARGE] at (image1.west) {\texttt{\textcolor{GHOST}{p}bench\textcolor{GHOST}{p}}};  
            \end{tikzpicture}
            \begin{tikzpicture}
              \node[inner sep=0] (image2) {\includegraphics[height=5cm]{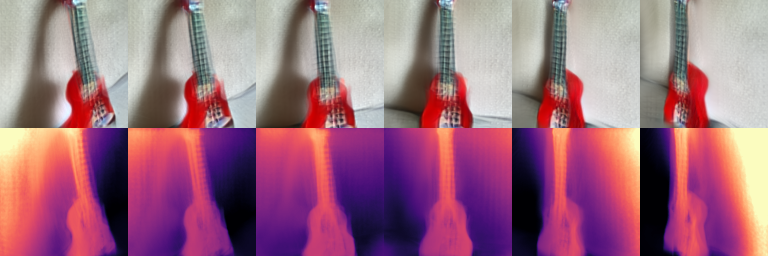}};
              \node[rotate=90, anchor=south, font=\LARGE] at (image2.west) {\texttt{guitar}};  
            \end{tikzpicture}%
        }
    \caption{MVImgNet}
    \label{subfig:gen-mvimgnet-app}
    \end{subfigure}
    
    \begin{subfigure}{\textwidth}
        \resizebox{\textwidth}{!}{%
            \begin{tikzpicture}
              \node[inner sep=0] (image1) {\includegraphics[height=5cm]{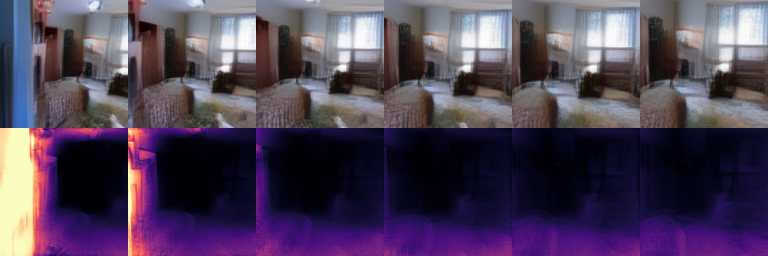}};
            \end{tikzpicture}
            \hspace{1em}
            \begin{tikzpicture}
              \node[inner sep=0] (image2) {\includegraphics[height=5cm]{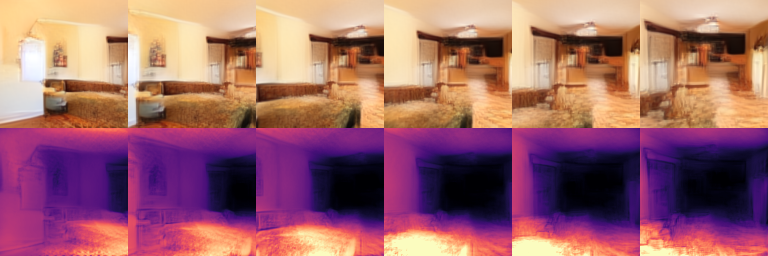}};
            \end{tikzpicture}%
        }
        \resizebox{\textwidth}{!}{%
            \begin{tikzpicture}
              \node[inner sep=0] (image1) {\includegraphics[height=5cm]{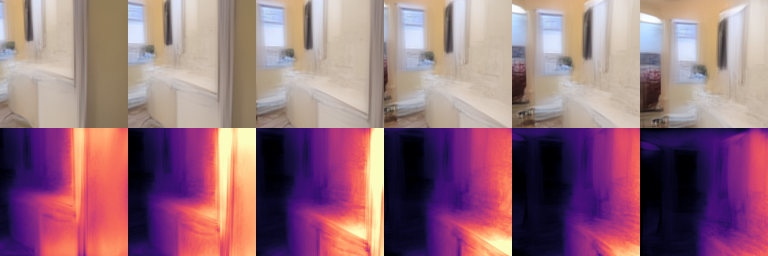}};
            \end{tikzpicture}
            \hspace{1em}
            \begin{tikzpicture}
              \node[inner sep=0] (image2) {\includegraphics[height=5cm]{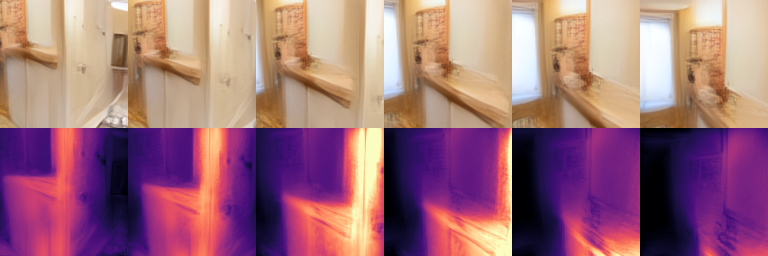}};
            \end{tikzpicture}%
        }
        \resizebox{\textwidth}{!}{%
            \begin{tikzpicture}
              \node[inner sep=0] (image1) {\includegraphics[height=5cm]{figures/gen/re10k/04db26572a791881_02.png}};
            \end{tikzpicture}
            \hspace{1em}
            \begin{tikzpicture}
              \node[inner sep=0] (image2) {\includegraphics[height=5cm]{figures/gen/re10k/0555b07fe6239b4a_01.png}};
            \end{tikzpicture}%
        }
    \caption{RealEstate10K}
    \label{subfig:gen-re10k-app}
    \end{subfigure}
\caption{Additional qualitative examples of class-conditional (MVImgNet) and unconditional generations (RealEstate10K) from our method. For each example, the top row shows six rendered views of the sampled 3D scene, while the bottom row shows the corresponding depths. }
\label{fig:gen-app}
\end{figure}

\begin{figure}[htbp]
\centering
\vspace{-8pt}
    \begin{subfigure}{\textwidth}
    \resizebox{\textwidth}{!}{%
            \begin{minipage}[c]{0.05\textwidth}
                \resizebox{\textwidth}{!}{%
                    \begin{tikzpicture}
                      \node[inner sep=0] (image1) {\includegraphics[height=2cm]{figures/1-im-recon/mvimgnet/ours/2a002f7d_input.png}};
                    \end{tikzpicture}
                }
            \end{minipage}
            \begin{minipage}[c]{0.25\textwidth}
                \resizebox{\textwidth}{!}{%
                    \begin{tikzpicture}
                      \node[inner sep=0] (image1) {\includegraphics[height=2cm, clip, trim=9.2cm 0 54.4cm 0]{figures/1-im-recon/mvimgnet/ours/2a002f7d_02.png}};
                    \end{tikzpicture}
                }
            \end{minipage}
            \begin{minipage}[c]{0.55\textwidth}
                \resizebox{\textwidth}{!}{%
                    \begin{tikzpicture}
                      \node[inner sep=0] (image1) {\includegraphics[height=2cm, clip, trim=31.8cm 0 0 0]{figures/1-im-recon/mvimgnet/ours/2a002f7d_02.png}};

                    \end{tikzpicture}
                }
                \vspace{0.1em}
                \resizebox{\textwidth}{!}{%
                    \begin{tikzpicture}
                      \node[inner sep=0] (image2) {\includegraphics[height=2cm, clip, trim=23.9cm 0 0 0]{figures/1-im-recon/mvimgnet/si/2a002f7d_00.png}};
                    \end{tikzpicture}
                }
            \end{minipage}
    }
    \resizebox{\textwidth}{!}{%
            \begin{minipage}[c]{0.05\textwidth}
                \resizebox{\textwidth}{!}{%
                    \begin{tikzpicture}
                      \node[inner sep=0] (image1) {\includegraphics[height=2cm]{figures/1-im-recon/mvimgnet/ours/260052d1_input.png}};
                    \end{tikzpicture}
                }
            \end{minipage}
            \begin{minipage}[c]{0.25\textwidth}
                \resizebox{\textwidth}{!}{%
                    \begin{tikzpicture}
                      \node[inner sep=0] (image1) {\includegraphics[height=2cm, clip, trim=9.2cm 0 54.4cm 0]{figures/1-im-recon/mvimgnet/ours/260052d1_00.png}};
                    \end{tikzpicture}
                }
            \end{minipage}
            \begin{minipage}[c]{0.55\textwidth}
                \resizebox{\textwidth}{!}{%
                    \begin{tikzpicture}
                      \node[inner sep=0] (image1) {\includegraphics[height=2cm, clip, trim=31.8cm 0 0 0]{figures/1-im-recon/mvimgnet/ours/260052d1_00.png}};

                    \end{tikzpicture}
                }
                \vspace{0.1em}
                \resizebox{\textwidth}{!}{%
                    \begin{tikzpicture}
                      \node[inner sep=0] (image2) {\includegraphics[height=2cm, clip, trim=23.9cm 0 0 0]{figures/1-im-recon/mvimgnet/si/260052d1_00.png}};
                    \end{tikzpicture}
                }
            \end{minipage}
    }
    \resizebox{\textwidth}{!}{%
            \begin{minipage}[c]{0.05\textwidth}
                \resizebox{\textwidth}{!}{%
                    \begin{tikzpicture}
                      \node[inner sep=0] (image1) {\includegraphics[height=2cm]{figures/1-im-recon/mvimgnet/ours/21005ead_input.png}};
                    \end{tikzpicture}
                }
            \end{minipage}
            \begin{minipage}[c]{0.25\textwidth}
                \resizebox{\textwidth}{!}{%
                    \begin{tikzpicture}
                      \node[inner sep=0] (image1) {\includegraphics[height=2cm, clip, trim=9.2cm 0 54.4cm 0]{figures/1-im-recon/mvimgnet/ours/21005ead_00.png}};
                    \end{tikzpicture}
                }
            \end{minipage}
            \begin{minipage}[c]{0.55\textwidth}
                \resizebox{\textwidth}{!}{%
                    \begin{tikzpicture}
                      \node[inner sep=0] (image1) {\includegraphics[height=2cm, clip, trim=31.8cm 0 0 0]{figures/1-im-recon/mvimgnet/ours/21005ead_00.png}};

                    \end{tikzpicture}
                }
                \vspace{0.1em}
                \resizebox{\textwidth}{!}{%
                    \begin{tikzpicture}
                      \node[inner sep=0] (image2) {\includegraphics[height=2cm, clip, trim=23.9cm 0 0 0]{figures/1-im-recon/mvimgnet/si/21005ead_00.png}};
                    \end{tikzpicture}
                }
            \end{minipage}
    }
    \resizebox{\textwidth}{!}{%
            \begin{minipage}[c]{0.05\textwidth}
                \resizebox{\textwidth}{!}{%
                    \begin{tikzpicture}
                      \node[inner sep=0] (image1) {\includegraphics[height=2cm]{figures/1-im-recon/mvimgnet/ours/030083be_input.png}};
                    \end{tikzpicture}
                }
            \end{minipage}
            \begin{minipage}[c]{0.25\textwidth}
                \resizebox{\textwidth}{!}{%
                    \begin{tikzpicture}
                      \node[inner sep=0] (image1) {\includegraphics[height=2cm, clip, trim=9.2cm 0 54.4cm 0]{figures/1-im-recon/mvimgnet/ours/030083be_02.png}};
                    \end{tikzpicture}
                }
            \end{minipage}
            \begin{minipage}[c]{0.55\textwidth}
                \resizebox{\textwidth}{!}{%
                    \begin{tikzpicture}
                      \node[inner sep=0] (image1) {\includegraphics[height=2cm, clip, trim=31.8cm 0 0 0]{figures/1-im-recon/mvimgnet/ours/030083be_02.png}};

                    \end{tikzpicture}
                }
                \vspace{0.1em}
                \resizebox{\textwidth}{!}{%
                    \begin{tikzpicture}
                      \node[inner sep=0] (image2) {\includegraphics[height=2cm, clip, trim=23.9cm 0 0 0]{figures/1-im-recon/mvimgnet/si/030083be_00.png}};
                    \end{tikzpicture}
                }
            \end{minipage}
    }
    \resizebox{\textwidth}{!}{%
            \begin{minipage}[c]{0.05\textwidth}
                \resizebox{\textwidth}{!}{%
                    \begin{tikzpicture}
                      \node[inner sep=0] (image1) {\includegraphics[height=2cm]{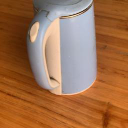}};
                    \end{tikzpicture}
                }
            \end{minipage}
            \begin{minipage}[c]{0.25\textwidth}
                \resizebox{\textwidth}{!}{%
                    \begin{tikzpicture}
                      \node[inner sep=0] (image1) {\includegraphics[height=2cm, clip, trim=9.2cm 0 54.4cm 0]{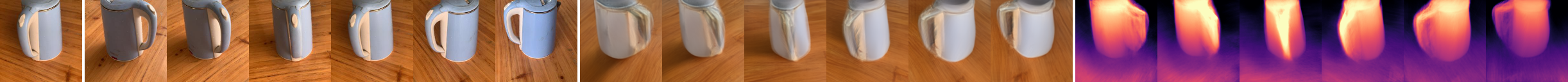}};
                    \end{tikzpicture}
                }
            \end{minipage}
            \begin{minipage}[c]{0.55\textwidth}
                \resizebox{\textwidth}{!}{%
                    \begin{tikzpicture}
                      \node[inner sep=0] (image1) {\includegraphics[height=2cm, clip, trim=31.8cm 0 0 0]{figures/1-im-recon/mvimgnet/ours/1c000dc6_03.png}};

                    \end{tikzpicture}
                }
                \vspace{0.1em}
                \resizebox{\textwidth}{!}{%
                    \begin{tikzpicture}
                      \node[inner sep=0] (image2) {\includegraphics[height=2cm, clip, trim=23.9cm 0 0 0]{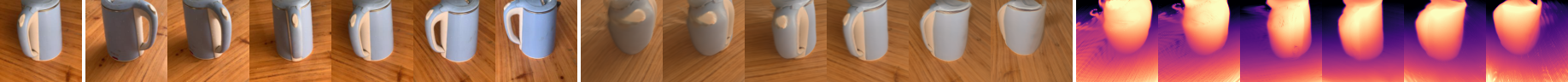}};
                    \end{tikzpicture}
                }
            \end{minipage}
    }
    \resizebox{\textwidth}{!}{%
            \begin{minipage}[c]{0.05\textwidth}
                \resizebox{\textwidth}{!}{%
                    \begin{tikzpicture}
                      \node[inner sep=0] (image1) {\includegraphics[height=2cm]{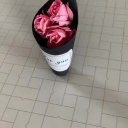}};
                    \end{tikzpicture}
                }
            \end{minipage}
            \begin{minipage}[c]{0.25\textwidth}
                \resizebox{\textwidth}{!}{%
                    \begin{tikzpicture}
                      \node[inner sep=0] (image1) {\includegraphics[height=2cm, clip, trim=9.2cm 0 54.4cm 0]{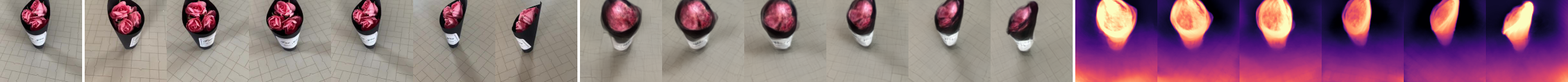}};
                    \end{tikzpicture}
                }
            \end{minipage}
            \begin{minipage}[c]{0.55\textwidth}
                \resizebox{\textwidth}{!}{%
                    \begin{tikzpicture}
                      \node[inner sep=0] (image1) {\includegraphics[height=2cm, clip, trim=31.8cm 0 0 0]{figures/1-im-recon/mvimgnet/ours/170114c3_02.png}};

                    \end{tikzpicture}
                }
                \vspace{0.1em}
                \resizebox{\textwidth}{!}{%
                    \begin{tikzpicture}
                      \node[inner sep=0] (image2) {\includegraphics[height=2cm, clip, trim=23.9cm 0 0 0]{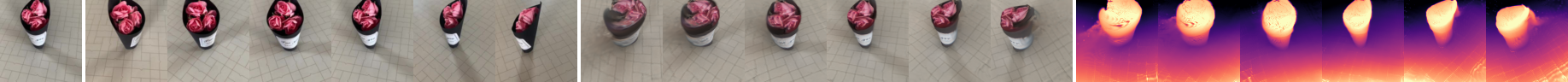}};
                    \end{tikzpicture}
                }
            \end{minipage}
    }
    \resizebox{\textwidth}{!}{%
            \begin{minipage}[c]{0.05\textwidth}
                \resizebox{\textwidth}{!}{%
                    \begin{tikzpicture}
                      \node[inner sep=0] (image1) {\includegraphics[height=2cm]{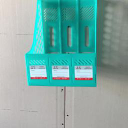}};
                    \end{tikzpicture}
                }
            \end{minipage}
            \begin{minipage}[c]{0.25\textwidth}
                \resizebox{\textwidth}{!}{%
                    \begin{tikzpicture}
                      \node[inner sep=0] (image1) {\includegraphics[height=2cm, clip, trim=9.2cm 0 54.4cm 0]{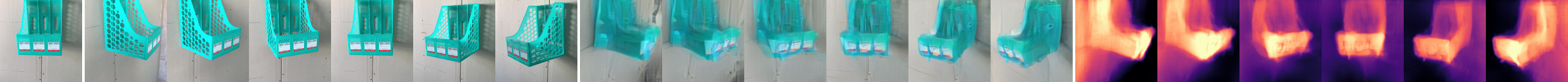}};
                    \end{tikzpicture}
                }
            \end{minipage}
            \begin{minipage}[c]{0.55\textwidth}
                \resizebox{\textwidth}{!}{%
                    \begin{tikzpicture}
                      \node[inner sep=0] (image1) {\includegraphics[height=2cm, clip, trim=31.8cm 0 0 0]{figures/1-im-recon/mvimgnet/ours/09008c53_03.png}};

                    \end{tikzpicture}
                }
                \vspace{0.1em}
                \resizebox{\textwidth}{!}{%
                    \begin{tikzpicture}
                      \node[inner sep=0] (image2) {\includegraphics[height=2cm, clip, trim=23.9cm 0 0 0]{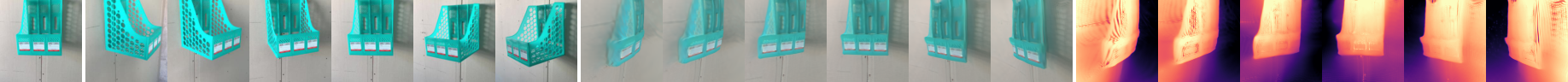}};
                    \end{tikzpicture}
                }
            \end{minipage}
    }
    \resizebox{\textwidth}{!}{%
            \begin{minipage}[c]{0.05\textwidth}
                \resizebox{\textwidth}{!}{%
                    \begin{tikzpicture}
                      \node[inner sep=0] (image1) {\includegraphics[height=2cm]{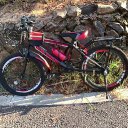}};
                    \end{tikzpicture}
                }
            \end{minipage}
            \begin{minipage}[c]{0.25\textwidth}
                \resizebox{\textwidth}{!}{%
                    \begin{tikzpicture}
                      \node[inner sep=0] (image1) {\includegraphics[height=2cm, clip, trim=9.2cm 0 54.4cm 0]{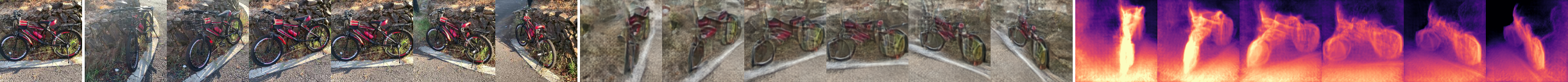}};
                    \end{tikzpicture}
                }
            \end{minipage}
            \begin{minipage}[c]{0.55\textwidth}
                \resizebox{\textwidth}{!}{%
                    \begin{tikzpicture}
                      \node[inner sep=0] (image1) {\includegraphics[height=2cm, clip, trim=31.8cm 0 0 0]{figures/1-im-recon/mvimgnet/ours/0100d5f4_02.png}};

                    \end{tikzpicture}
                }
                \vspace{0.1em}
                \resizebox{\textwidth}{!}{%
                    \begin{tikzpicture}
                      \node[inner sep=0] (image2) {\includegraphics[height=2cm, clip, trim=23.9cm 0 0 0]{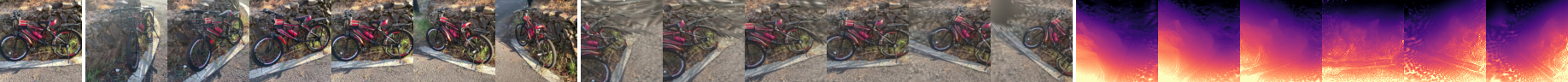}};
                    \end{tikzpicture}
                }
            \end{minipage}
    }
    \resizebox{\textwidth}{!}{%
            \begin{minipage}[c]{0.05\textwidth}
                \resizebox{\textwidth}{!}{%
                    \begin{tikzpicture}
                      \node[inner sep=0] (image1) {\includegraphics[height=2cm]{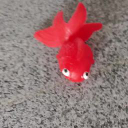}};
                    \end{tikzpicture}
                }
            \end{minipage}
            \begin{minipage}[c]{0.25\textwidth}
                \resizebox{\textwidth}{!}{%
                    \begin{tikzpicture}
                      \node[inner sep=0] (image1) {\includegraphics[height=2cm, clip, trim=9.2cm 0 54.4cm 0]{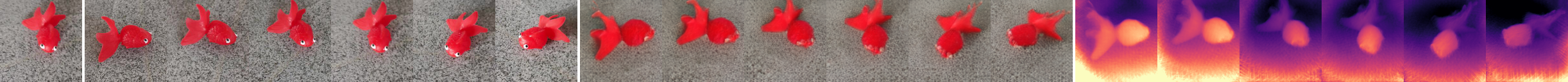}};
                    \end{tikzpicture}
                }
            \end{minipage}
            \begin{minipage}[c]{0.55\textwidth}
                \resizebox{\textwidth}{!}{%
                    \begin{tikzpicture}
                      \node[inner sep=0] (image1) {\includegraphics[height=2cm, clip, trim=31.8cm 0 0 0]{figures/1-im-recon/mvimgnet/ours/28008dc9_02.png}};

                    \end{tikzpicture}
                }
                \vspace{0.1em}
                \resizebox{\textwidth}{!}{%
                    \begin{tikzpicture}
                      \node[inner sep=0] (image2) {\includegraphics[height=2cm, clip, trim=23.9cm 0 0 0]{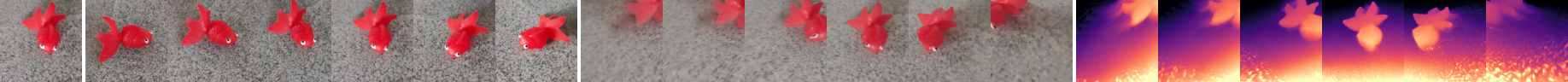}};
                    \end{tikzpicture}
                }
            \end{minipage}
    }
    \caption{MVImgNet}
    \label{subfig:1-im-recon-mvimgnet-app}
    \end{subfigure}

    \begin{subfigure}{\textwidth}
    \resizebox{\textwidth}{!}{%
            \begin{minipage}[c]{0.05\textwidth}
                \resizebox{\textwidth}{!}{%
                    \begin{tikzpicture}
                      \node[inner sep=0] (image1) {\includegraphics[height=2cm]{figures/1-im-recon/ra10k/ours/004dd4b46a06e5be_input.png}};
                    \end{tikzpicture}
                }
            \end{minipage}
            \begin{minipage}[c]{0.25\textwidth}
                \resizebox{\textwidth}{!}{%
                    \begin{tikzpicture}
                      \node[inner sep=0] (image1) {\includegraphics[height=2cm, clip, trim=9.2cm 0 54.4cm 0]{figures/1-im-recon/ra10k/ours/004dd4b46a06e5be_02.png}};
                    \end{tikzpicture}
                }
            \end{minipage}
            \begin{minipage}[c]{0.55\textwidth}
                \resizebox{\textwidth}{!}{%
                    \begin{tikzpicture}
                      \node[inner sep=0] (image1) {\includegraphics[height=2cm, clip, trim=31.8cm 0 0 0]{figures/1-im-recon/ra10k/ours/004dd4b46a06e5be_02.png}};

                    \end{tikzpicture}
                }
                \vspace{0.1em}
                \resizebox{\textwidth}{!}{%
                    \begin{tikzpicture}
                      \node[inner sep=0] (image2) {\includegraphics[height=2cm, clip, trim=23.9cm 0 0 0]{figures/1-im-recon/ra10k/si/004dd4b46a06e5be_00.png}};
                    \end{tikzpicture}
                }
            \end{minipage}
    }
    \resizebox{\textwidth}{!}{%
            \begin{minipage}[c]{0.05\textwidth}
                \resizebox{\textwidth}{!}{%
                    \begin{tikzpicture}
                      \node[inner sep=0] (image1) {\includegraphics[height=2cm]{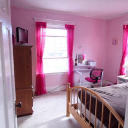}};
                    \end{tikzpicture}
                }
            \end{minipage}
            \begin{minipage}[c]{0.25\textwidth}
                \resizebox{\textwidth}{!}{%
                    \begin{tikzpicture}
                      \node[inner sep=0] (image1) {\includegraphics[height=2cm, clip, trim=9.2cm 0 54.4cm 0]{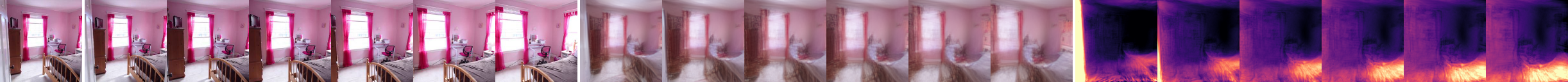}};
                    \end{tikzpicture}
                }
            \end{minipage}
            \begin{minipage}[c]{0.55\textwidth}
                \resizebox{\textwidth}{!}{%
                    \begin{tikzpicture}
                      \node[inner sep=0] (image1) {\includegraphics[height=2cm, clip, trim=31.8cm 0 0 0]{figures/1-im-recon/ra10k/ours/0af60a9ffd747a1c_03.png}};

                    \end{tikzpicture}
                }
                \vspace{0.1em}
                \resizebox{\textwidth}{!}{%
                    \begin{tikzpicture}
                      \node[inner sep=0] (image2) {\includegraphics[height=2cm, clip, trim=23.9cm 0 0 0]{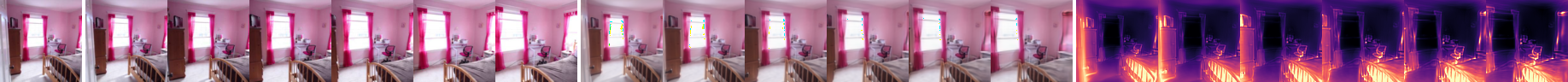}};
                    \end{tikzpicture}
                }
            \end{minipage}
    }
    \resizebox{\textwidth}{!}{%
            \begin{minipage}[c]{0.05\textwidth}
                \resizebox{\textwidth}{!}{%
                    \begin{tikzpicture}
                      \node[inner sep=0] (image1) {\includegraphics[height=2cm]{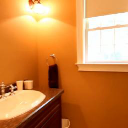}};
                    \end{tikzpicture}
                }
            \end{minipage}
            \begin{minipage}[c]{0.25\textwidth}
                \resizebox{\textwidth}{!}{%
                    \begin{tikzpicture}
                      \node[inner sep=0] (image1) {\includegraphics[height=2cm, clip, trim=9.2cm 0 54.4cm 0]{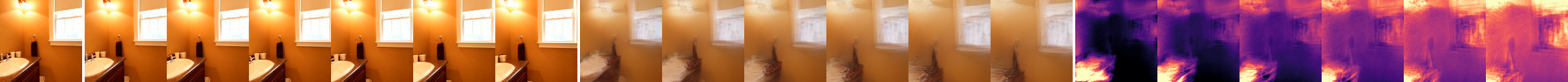}};
                    \end{tikzpicture}
                }
            \end{minipage}
            \begin{minipage}[c]{0.55\textwidth}
                \resizebox{\textwidth}{!}{%
                    \begin{tikzpicture}
                      \node[inner sep=0] (image1) {\includegraphics[height=2cm, clip, trim=31.8cm 0 0 0]{figures/1-im-recon/ra10k/ours/adb7ac21aa811ffc_01.png}};

                    \end{tikzpicture}
                }
                \vspace{0.1em}
                \resizebox{\textwidth}{!}{%
                    \begin{tikzpicture}
                      \node[inner sep=0] (image2) {\includegraphics[height=2cm, clip, trim=23.9cm 0 0 0]{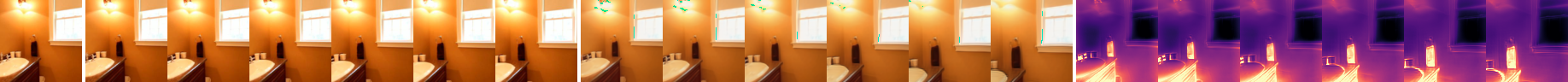}};
                    \end{tikzpicture}
                }
            \end{minipage}
    }
    \resizebox{\textwidth}{!}{%
            \begin{minipage}[c]{0.05\textwidth}
                \resizebox{\textwidth}{!}{%
                    \begin{tikzpicture}
                      \node[inner sep=0] (image1) {\includegraphics[height=2cm]{figures/1-im-recon/ra10k/ours/03ac738af49c7596_input.png}};
                    \end{tikzpicture}
                }
            \end{minipage}
            \begin{minipage}[c]{0.25\textwidth}
                \resizebox{\textwidth}{!}{%
                    \begin{tikzpicture}
                      \node[inner sep=0] (image1) {\includegraphics[height=2cm, clip, trim=9.2cm 0 54.4cm 0]{figures/1-im-recon/ra10k/ours/03ac738af49c7596_03.png}};
                    \end{tikzpicture}
                }
            \end{minipage}
            \begin{minipage}[c]{0.55\textwidth}
                \resizebox{\textwidth}{!}{%
                    \begin{tikzpicture}
                      \node[inner sep=0] (image1) {\includegraphics[height=2cm, clip, trim=31.8cm 0 0 0]{figures/1-im-recon/ra10k/ours/03ac738af49c7596_03.png}};

                    \end{tikzpicture}
                }
                \vspace{0.1em}
                \resizebox{\textwidth}{!}{%
                    \begin{tikzpicture}
                      \node[inner sep=0] (image2) {\includegraphics[height=2cm, clip, trim=23.9cm 0 0 0]{figures/1-im-recon/ra10k/si/03ac738af49c7596_00.png}};
                    \end{tikzpicture}
                }
            \end{minipage}
    }
    \caption{RealEstate10k}
    \label{subfig:1-im-recon-ra10k-app}
    \end{subfigure}
 \caption{Additional qualitative comparison of 3D reconstruction from a single image between our model (upper row of each pair) and SplatterImage (lower row of each pair) on MVImgNet (a) and RealEstate10k (b). The first column shows the input (conditioning) image, the second displays the ground truth images, and the third and fourth columns display the predicted frames and depths, respectively.}
 \label{fig:1-im-recon-app}
\end{figure}

\begin{figure}[htbp]
    \centering
    \resizebox{0.9\textwidth}{!}{%
            \begin{minipage}[c]{0.12\textwidth}
                \resizebox{\textwidth}{!}{%
                    \begin{tikzpicture}
                      \node[inner sep=0] (image1) {\includegraphics[height=2cm, clip, trim=0 0 4.6cm  0]{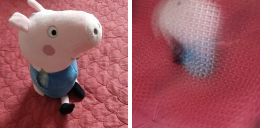}};
                    \end{tikzpicture}
                }
            \end{minipage}
            \begin{minipage}[c]{0.68\textwidth}
                \resizebox{\textwidth}{!}{%
                    \begin{tikzpicture}
                      \node[inner sep=0] (image1) {\includegraphics[height=2cm, clip, trim=13.7cm 0 9cm  0]{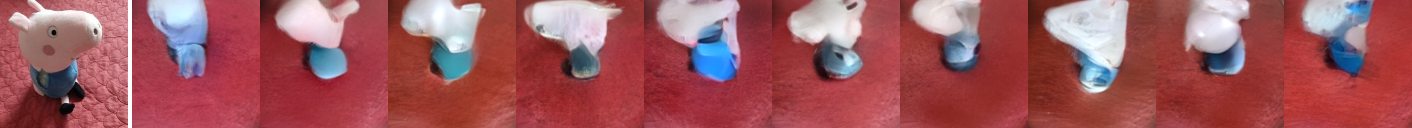}};
                    \end{tikzpicture}
                }
            \end{minipage}
            \begin{minipage}[c]{0.12\textwidth}
                \resizebox{\textwidth}{!}{%
                    \begin{tikzpicture}
                      \node[inner sep=0] (image1) {\includegraphics[height=2cm, clip, trim=4.6cm 0 0 0]{figures/back-diversity/det/00014a5f_backs.png}};
                    \end{tikzpicture}
                    
                }
            \end{minipage}
    }
    \resizebox{0.9\textwidth}{!}{%
            \begin{minipage}[c]{0.12\textwidth}
                \resizebox{\textwidth}{!}{%
                    \begin{tikzpicture}
                      \node[inner sep=0] (image1) {\includegraphics[height=2cm, clip, trim=0 0 4.6cm  0]{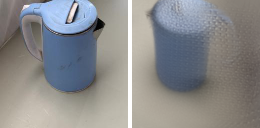}};
                    \end{tikzpicture}
                }
            \end{minipage}
            \begin{minipage}[c]{0.68\textwidth}
                \resizebox{\textwidth}{!}{%
                    \begin{tikzpicture}
                      \node[inner sep=0] (image1) {\includegraphics[height=2cm, clip, trim=13.7cm 0 9cm  0]{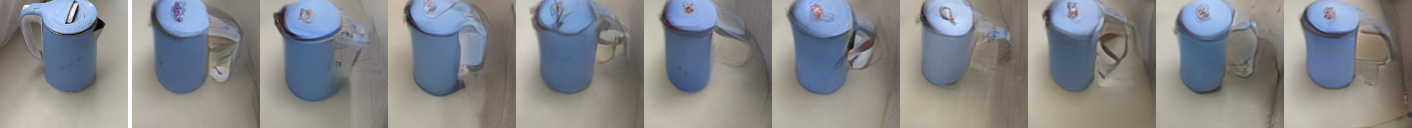}};
                    \end{tikzpicture}
                }
            \end{minipage}
            \begin{minipage}[c]{0.12\textwidth}
                \resizebox{\textwidth}{!}{%
                    \begin{tikzpicture}
                      \node[inner sep=0] (image1) {\includegraphics[height=2cm, clip, trim=4.6cm 0 0 0]{figures/back-diversity/det/1c008fee_backs.png}};
                    \end{tikzpicture}
                }
            \end{minipage}
    }
    \resizebox{0.9\textwidth}{!}{%
            \begin{minipage}[c]{0.12\textwidth}
                \resizebox{\textwidth}{!}{%
                    \begin{tikzpicture}
                      \node[inner sep=0] (image1) {\includegraphics[height=2cm, clip, trim=0 0 4.6cm  0]{figures/back-diversity/det/1c012ff3_backs.png}};
                    \end{tikzpicture}
                }
            \end{minipage}
            \begin{minipage}[c]{0.68\textwidth}
                \resizebox{\textwidth}{!}{%
                    \begin{tikzpicture}
                      \node[inner sep=0] (image1) {\includegraphics[height=2cm, clip, trim=13.7cm 0 9cm  0]{figures/back-diversity/gen/1c012ff3_backs.png}};
                    \end{tikzpicture}
                }
            \end{minipage}
            \begin{minipage}[c]{0.12\textwidth}
                \resizebox{\textwidth}{!}{%
                    \begin{tikzpicture}
                      \node[inner sep=0] (image1) {\includegraphics[height=2cm, clip, trim=4.6cm 0 0 0]{figures/back-diversity/det/1c012ff3_backs.png}};
                    \end{tikzpicture}
                }
            \end{minipage}       
    }
    \resizebox{0.9\textwidth}{!}{%
            \begin{minipage}[c]{0.12\textwidth}
                \resizebox{\textwidth}{!}{%
                    \begin{tikzpicture}
                      \node[inner sep=0] (image1) {\includegraphics[height=2cm, clip, trim=0 0 4.6cm  0]{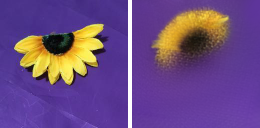}};
                    \end{tikzpicture}
                }
            \end{minipage}
            \begin{minipage}[c]{0.68\textwidth}
                \resizebox{\textwidth}{!}{%
                    \begin{tikzpicture}
                      \node[inner sep=0] (image1) {\includegraphics[height=2cm, clip, trim=13.7cm 0 9cm  0]{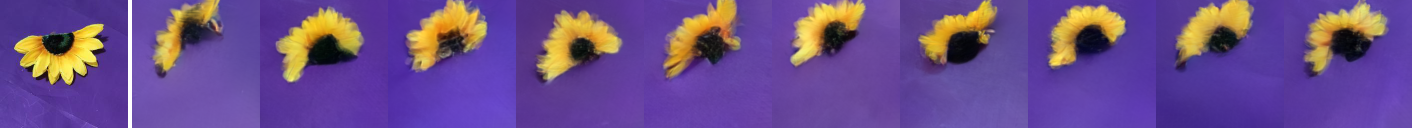}};
                    \end{tikzpicture}
                }
            \end{minipage}
            \begin{minipage}[c]{0.12\textwidth}
                \resizebox{\textwidth}{!}{%
                    \begin{tikzpicture}
                      \node[inner sep=0] (image1) {\includegraphics[height=2cm, clip, trim=4.6cm 0 0 0]{figures/back-diversity/det/1d011add_backs.png}};
                    \end{tikzpicture}
                }
            \end{minipage}            
    }
    \resizebox{0.9\textwidth}{!}{%
            \begin{minipage}[c]{0.12\textwidth}
                \resizebox{\textwidth}{!}{%
                    \begin{tikzpicture}
                      \node[inner sep=0] (image1) {\includegraphics[height=2cm, clip, trim=0 0 4.6cm  0]{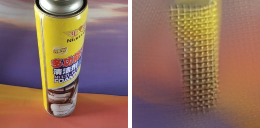}};
                    \end{tikzpicture}
                }
            \end{minipage}
            \begin{minipage}[c]{0.68\textwidth}
                \resizebox{\textwidth}{!}{%
                    \begin{tikzpicture}
                      \node[inner sep=0] (image1) {\includegraphics[height=2cm, clip, trim=13.7cm 0 9cm  0]{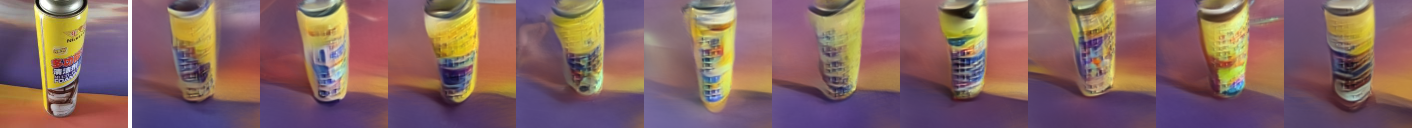}};
                    \end{tikzpicture}
                }
            \end{minipage}
            \begin{minipage}[c]{0.12\textwidth}
                \resizebox{\textwidth}{!}{%
                    \begin{tikzpicture}
                      \node[inner sep=0] (image1) {\includegraphics[height=2cm, clip, trim=4.6cm 0 0 0]{figures/back-diversity/det/1f00991e_backs.png}};
                    \end{tikzpicture}
                }
            \end{minipage}
    }
    \resizebox{0.9\textwidth}{!}{%
            \begin{minipage}[c]{0.12\textwidth}
                \resizebox{\textwidth}{!}{%
                    \begin{tikzpicture}
                      \node[inner sep=0] (image1) {\includegraphics[height=2cm, clip, trim=0 0 4.6cm  0]{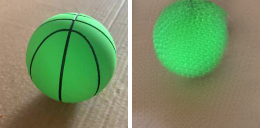}};
                    \end{tikzpicture}
                }
            \end{minipage}
            \begin{minipage}[c]{0.68\textwidth}
                \resizebox{\textwidth}{!}{%
                    \begin{tikzpicture}
                      \node[inner sep=0] (image1) {\includegraphics[height=2cm, clip, trim=13.7cm 0 9cm  0]{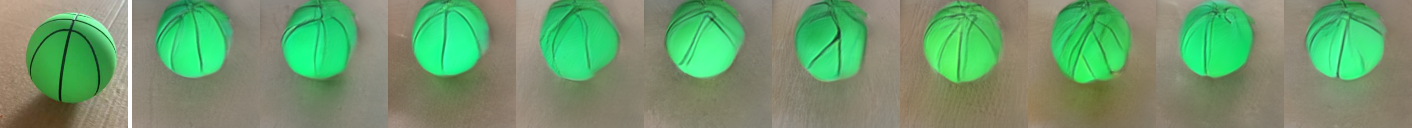}};
                    \end{tikzpicture}
                }
            \end{minipage}
            \begin{minipage}[c]{0.12\textwidth}
                \resizebox{\textwidth}{!}{%
                    \begin{tikzpicture}
                      \node[inner sep=0] (image1) {\includegraphics[height=2cm, clip, trim=4.6cm 0 0 0]{figures/back-diversity/det/2a011358_backs.png}};
                    \end{tikzpicture}
                }
            \end{minipage}
    }
    \resizebox{0.9\textwidth}{!}{%
            \begin{minipage}[c]{0.12\textwidth}
                \resizebox{\textwidth}{!}{%
                    \begin{tikzpicture}
                      \node[inner sep=0] (image1) {\includegraphics[height=2cm, clip, trim=0 0 4.6cm  0]{figures/back-diversity/det/1800ce8b_backs.png}};
                    \end{tikzpicture}
                }
            \end{minipage}
            \begin{minipage}[c]{0.68\textwidth}
                \resizebox{\textwidth}{!}{%
                    \begin{tikzpicture}
                      \node[inner sep=0] (image1) {\includegraphics[height=2cm, clip, trim=13.7cm 0 9cm  0]{figures/back-diversity/gen/1800ce8b_backs.png}};
                    \end{tikzpicture}
                }
            \end{minipage}
            \begin{minipage}[c]{0.12\textwidth}
                \resizebox{\textwidth}{!}{%
                    \begin{tikzpicture}
                      \node[inner sep=0] (image1) {\includegraphics[height=2cm, clip, trim=4.6cm 0 0 0]{figures/back-diversity/det/1800ce8b_backs.png}};
                    \end{tikzpicture}
                }
            \end{minipage}
    }
    \resizebox{0.9\textwidth}{!}{%
            \begin{minipage}[c]{0.12\textwidth}
                \resizebox{\textwidth}{!}{%
                    \begin{tikzpicture}
                      \node[inner sep=0] (image1) {\includegraphics[height=2cm, clip, trim=0 0 4.6cm  0]{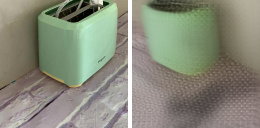}};
                    \end{tikzpicture}
                }
            \end{minipage}
            \begin{minipage}[c]{0.68\textwidth}
                \resizebox{\textwidth}{!}{%
                    \begin{tikzpicture}
                      \node[inner sep=0] (image1) {\includegraphics[height=2cm, clip, trim=13.7cm 0 9cm  0]{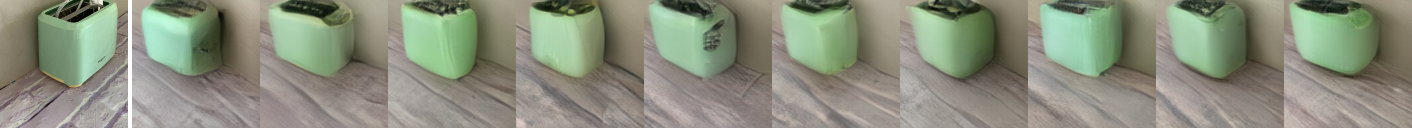}};
                    \end{tikzpicture}
                }
            \end{minipage}
            \begin{minipage}[c]{0.12\textwidth}
                \resizebox{\textwidth}{!}{%
                    \begin{tikzpicture}
                      \node[inner sep=0] (image1) {\includegraphics[height=2cm, clip, trim=4.6cm 0 0 0]{figures/back-diversity/det/040121f0_backs.png}};
                    \end{tikzpicture}
                }
            \end{minipage}
    }
    \caption{Additional qualitative results for 3D reconstruction from a single image showing diversity of sampled 3D assets. Given a single view as input (first column), our model samples 3D scenes from the posterior distribution (center columns). In contrast to a deterministic baseline, which outputs averaged blurry solution (last column), our model samples diverse plausible back-views}
    \label{fig:diverse-backs-app}
\end{figure}

Figures~\ref{fig:gen-app}--\ref{fig:diverse-backs-app} give additional qualitative results, following the same protocol as the corresponding figures in the paper. Please refer to the respective captions for details.

\section{Compute Requirements}
\label{app:compute}

Our final models each trained for 24 GPU-days on a local cluster, on nodes with $4\times$ NVIDIA A5000 GPUs (24GB VRAM) and single 16-core CPU with 128GB RAM; we used either two or four GPUs per run.
SplatterImage trained for approximately 4 GPU-days on the same hardware, using a single GPU per run.
The total compute for the project (including preliminary runs and hyperparameter sweeps) is estimated at 2500 GPU-days.

\begin{table}
    \centering
    \caption{Quality of denoised views, compared with and held-out views placed between them. Comparable values indicate that although denoising occurs (and splats are supported on) a subset of views, the 3D shape also re-renders accurately to other views }
    \label{tab:diff-vs-heldout-results}
    \begin{tabular}{@{} l cccc cccc @{}}
    \toprule
      & \multicolumn{3}{c}{\textbf{diffused}} & \multicolumn{3}{c@{}}{\textbf{held-out}} \\
      \cmidrule{2-4}\cmidrule(l){5-7}
      & FID $\downarrow$ & PSNR $\uparrow$ & LPIPS $\downarrow$ & FID $\downarrow$ & PSNR $\uparrow$ & LPIPS $\downarrow$ \\
      \midrule
      MVImgNet   & 22.8 & 20.3 & 0.332 & 23.6 & 20.9 & 0.318 \\
      RealEstate10K  & 29.3 & 16.4 & 0.457 & 29.9 & 16.5 & 0.453 \\
      \bottomrule
    \end{tabular}
\end{table}

\begin{figure}[htbp]
    \centering
    \begin{subfigure}{\textwidth}
    \resizebox{\textwidth}{!}{%
            \begin{minipage}[c]{0.2\textwidth}
                \resizebox{\textwidth}{!}{%
                    \begin{tikzpicture}
                      \node[inner sep=0] (image1) {\includegraphics[height=2cm]{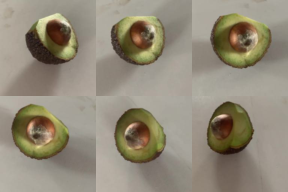}};
                    \end{tikzpicture}
                }
            \end{minipage}
            \begin{minipage}[c]{0.39\textwidth}
                \resizebox{\textwidth}{!}{%
                    \begin{tikzpicture}
                      \node[inner sep=0] (image1) {\includegraphics[height=2cm]{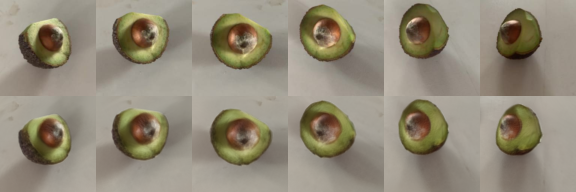}};
                    \end{tikzpicture}
                }
            \end{minipage}
            \begin{minipage}[c]{0.38\textwidth}
                \resizebox{\textwidth}{!}{%
                    \begin{tikzpicture}
                      \node[inner sep=0] (image1) {\includegraphics[height=2cm, clip, trim=61.3cm 0 0 0]{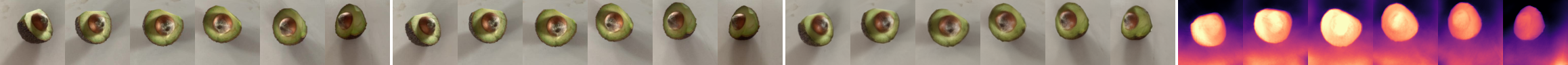}};
                    \end{tikzpicture}
                }
            \end{minipage}
    }
    \resizebox{\textwidth}{!}{%
            \begin{minipage}[c]{0.2\textwidth}
                \resizebox{\textwidth}{!}{%
                    \begin{tikzpicture}
                      \node[inner sep=0] (image1) {\includegraphics[height=2cm]{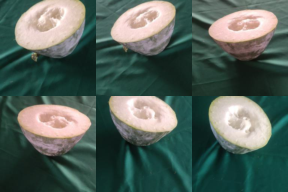}};
                    \end{tikzpicture}
                }
            \end{minipage}
            \begin{minipage}[c]{0.39\textwidth}
                \resizebox{\textwidth}{!}{%
                    \begin{tikzpicture}
                      \node[inner sep=0] (image1) {\includegraphics[height=2cm]{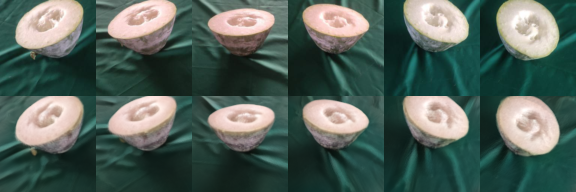}};
                    \end{tikzpicture}
                }
            \end{minipage}
            \begin{minipage}[c]{0.38\textwidth}
                \resizebox{\textwidth}{!}{%
                    \begin{tikzpicture}
                      \node[inner sep=0] (image1) {\includegraphics[height=2cm, clip, trim=61.3cm 0 0 0]{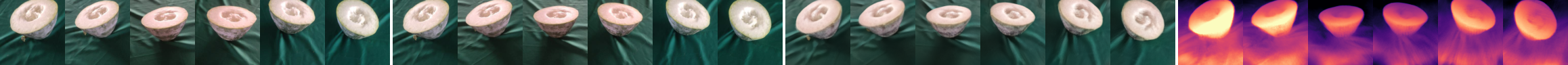}};
                    \end{tikzpicture}
                }
            \end{minipage}
    }
    \resizebox{\textwidth}{!}{%
            \begin{minipage}[c]{0.2\textwidth}
                \resizebox{\textwidth}{!}{%
                    \begin{tikzpicture}
                      \node[inner sep=0] (image1) {\includegraphics[height=2cm]{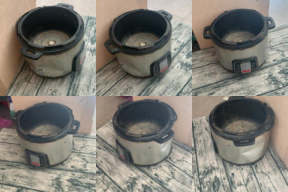}};
                    \end{tikzpicture}
                }
            \end{minipage}
            \begin{minipage}[c]{0.39\textwidth}
                \resizebox{\textwidth}{!}{%
                    \begin{tikzpicture}
                      \node[inner sep=0] (image1) {\includegraphics[height=2cm]{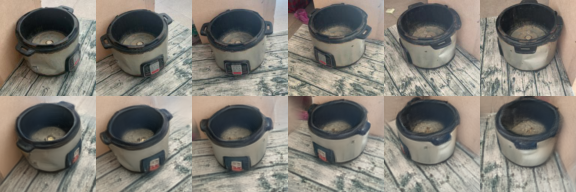}};
                    \end{tikzpicture}
                }
            \end{minipage}
            \begin{minipage}[c]{0.38\textwidth}
                \resizebox{\textwidth}{!}{%
                    \begin{tikzpicture}
                      \node[inner sep=0] (image1) {\includegraphics[height=2cm, clip, trim=61.3cm 0 0 0]{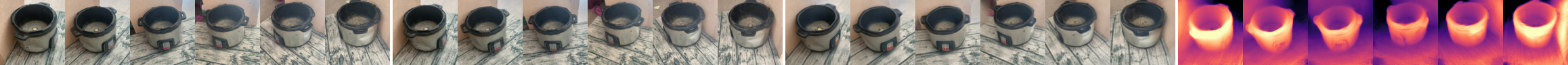}};
                    \end{tikzpicture}
                }
            \end{minipage}
    }
    \resizebox{\textwidth}{!}{%
            \begin{minipage}[c]{0.2\textwidth}
                \resizebox{\textwidth}{!}{%
                    \begin{tikzpicture}
                      \node[inner sep=0] (image1) {\includegraphics[height=2cm]{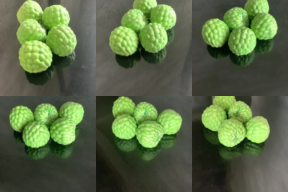}};
                    \end{tikzpicture}
                }
            \end{minipage}
            \begin{minipage}[c]{0.39\textwidth}
                \resizebox{\textwidth}{!}{%
                    \begin{tikzpicture}
                      \node[inner sep=0] (image1) {\includegraphics[height=2cm]{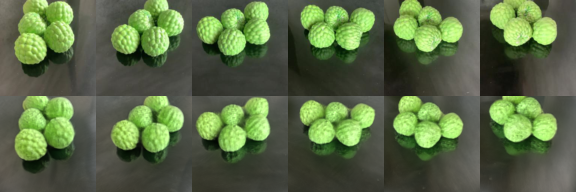}};
                    \end{tikzpicture}
                }
            \end{minipage}
            \begin{minipage}[c]{0.38\textwidth}
                \resizebox{\textwidth}{!}{%
                    \begin{tikzpicture}
                      \node[inner sep=0] (image1) {\includegraphics[height=2cm, clip, trim=61.3cm 0 0 0]{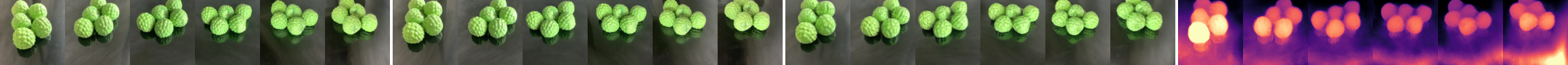}};
                    \end{tikzpicture}
                }
            \end{minipage}
    }
    \resizebox{\textwidth}{!}{%
            \begin{minipage}[c]{0.2\textwidth}
                \resizebox{\textwidth}{!}{%
                    \begin{tikzpicture}
                      \node[inner sep=0] (image1) {\includegraphics[height=2cm]{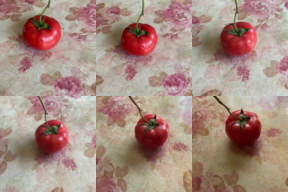}};
                    \end{tikzpicture}
                }
            \end{minipage}
            \begin{minipage}[c]{0.39\textwidth}
                \resizebox{\textwidth}{!}{%
                    \begin{tikzpicture}
                      \node[inner sep=0] (image1) {\includegraphics[height=2cm]{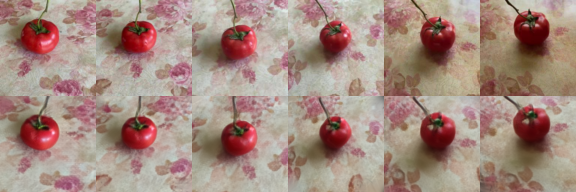}};
                    \end{tikzpicture}
                }
            \end{minipage}
            \begin{minipage}[c]{0.38\textwidth}
                \resizebox{\textwidth}{!}{%
                    \begin{tikzpicture}
                      \node[inner sep=0] (image1) {\includegraphics[height=2cm, clip, trim=61.3cm 0 0 0]{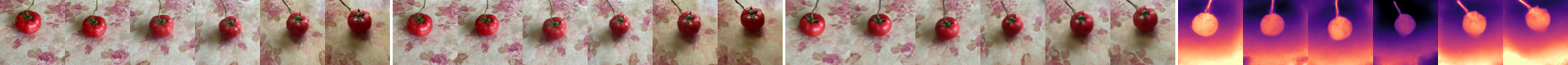}};
                    \end{tikzpicture}
                }
            \end{minipage}
    }
    \resizebox{\textwidth}{!}{%
            \begin{minipage}[c]{0.2\textwidth}
                \resizebox{\textwidth}{!}{%
                    \begin{tikzpicture}
                      \node[inner sep=0] (image1) {\includegraphics[height=2cm]{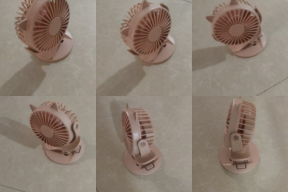}};
                    \end{tikzpicture}
                }
            \end{minipage}
            \begin{minipage}[c]{0.39\textwidth}
                \resizebox{\textwidth}{!}{%
                    \begin{tikzpicture}
                      \node[inner sep=0] (image1) {\includegraphics[height=2cm]{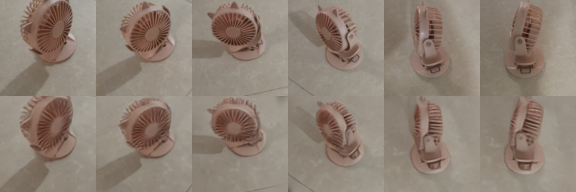}};
                    \end{tikzpicture}
                }
            \end{minipage}
            \begin{minipage}[c]{0.38\textwidth}
                \resizebox{\textwidth}{!}{%
                    \begin{tikzpicture}
                      \node[inner sep=0] (image1) {\includegraphics[height=2cm, clip, trim=61.3cm 0 0 0]{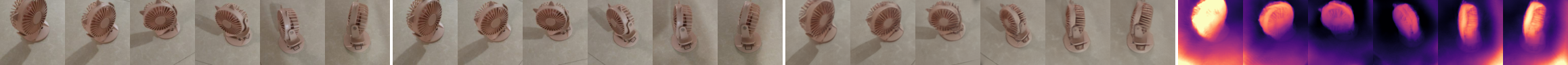}};
                    \end{tikzpicture}
                }
            \end{minipage}
    }
    \resizebox{\textwidth}{!}{%
            \begin{minipage}[c]{0.2\textwidth}
                \resizebox{\textwidth}{!}{%
                    \begin{tikzpicture}
                      \node[inner sep=0] (image1) {\includegraphics[height=2cm]{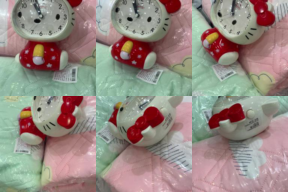}};
                    \end{tikzpicture}
                }
            \end{minipage}
            \begin{minipage}[c]{0.39\textwidth}
                \resizebox{\textwidth}{!}{%
                    \begin{tikzpicture}
                      \node[inner sep=0] (image1) {\includegraphics[height=2cm]{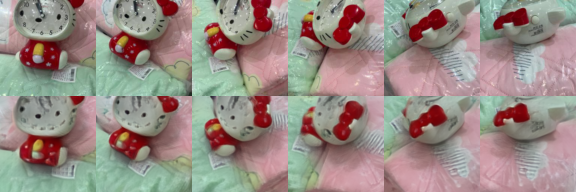}};
                    \end{tikzpicture}
                }
            \end{minipage}
            \begin{minipage}[c]{0.38\textwidth}
                \resizebox{\textwidth}{!}{%
                    \begin{tikzpicture}
                      \node[inner sep=0] (image1) {\includegraphics[height=2cm, clip, trim=61.3cm 0 0 0]{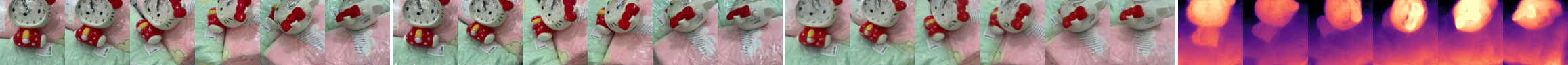}};
                    \end{tikzpicture}
                }
            \end{minipage}
    }
    \resizebox{\textwidth}{!}{%
            \begin{minipage}[c]{0.2\textwidth}
                \resizebox{\textwidth}{!}{%
                    \begin{tikzpicture}
                      \node[inner sep=0] (image1) {\includegraphics[height=2cm]{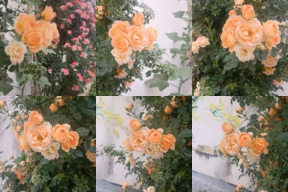}};
                    \end{tikzpicture}
                }
            \end{minipage}
            \begin{minipage}[c]{0.39\textwidth}
                \resizebox{\textwidth}{!}{%
                    \begin{tikzpicture}
                      \node[inner sep=0] (image1) {\includegraphics[height=2cm]{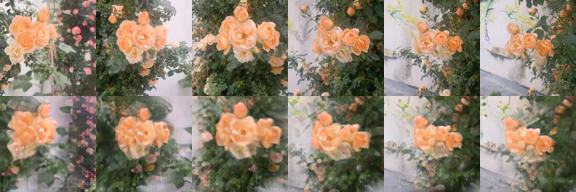}};
                    \end{tikzpicture}
                }
            \end{minipage}
            \begin{minipage}[c]{0.38\textwidth}
                \resizebox{\textwidth}{!}{%
                    \begin{tikzpicture}
                      \node[inner sep=0] (image1) {\includegraphics[height=2cm, clip, trim=61.3cm 0 0 0]{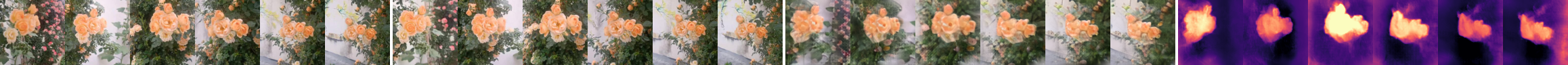}};
                    \end{tikzpicture}
                }
            \end{minipage}
    }
    \caption{MVImageNet}
    \label{subfig:6-im-recon-mvimgnet-app}
    \end{subfigure}
    \begin{subfigure}{\textwidth}
    \resizebox{\textwidth}{!}{%
            \begin{minipage}[c]{0.2\textwidth}
                \resizebox{\textwidth}{!}{%
                    \begin{tikzpicture}
                      \node[inner sep=0] (image1) {\includegraphics[height=2cm]{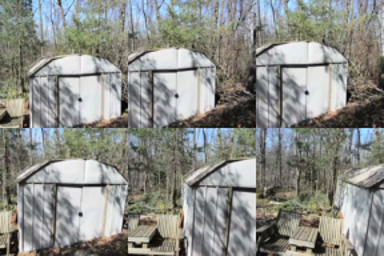}};

                    \end{tikzpicture}
                }
            \end{minipage}
            \begin{minipage}[c]{0.39\textwidth}
                \resizebox{\textwidth}{!}{%
                    \begin{tikzpicture}
                      \node[inner sep=0] (image1) {\includegraphics[height=2cm]{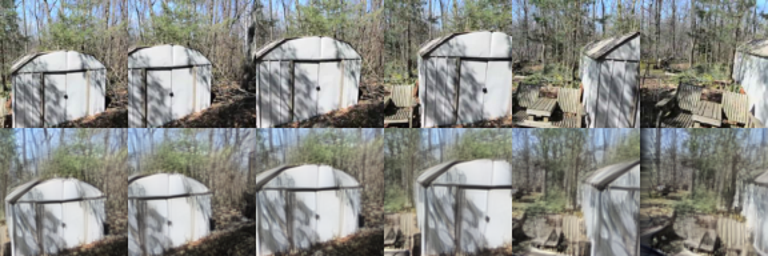}};

                    \end{tikzpicture}
                }
            \end{minipage}
            \begin{minipage}[c]{0.38\textwidth}
                \resizebox{\textwidth}{!}{%
                    \begin{tikzpicture}
                      \node[inner sep=0] (image1) {\includegraphics[height=2cm, clip, trim=61.3cm 0 0 0]{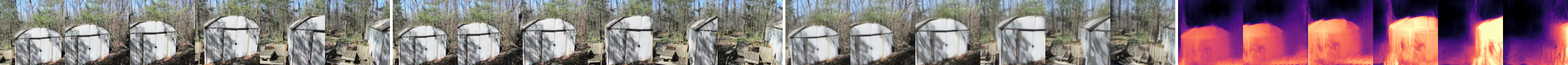}};

                    \end{tikzpicture}
                }
            \end{minipage}
    }
    \resizebox{\textwidth}{!}{%
            \begin{minipage}[c]{0.2\textwidth}
                \resizebox{\textwidth}{!}{%
                    \begin{tikzpicture}
                      \node[inner sep=0] (image1) {\includegraphics[height=2cm]{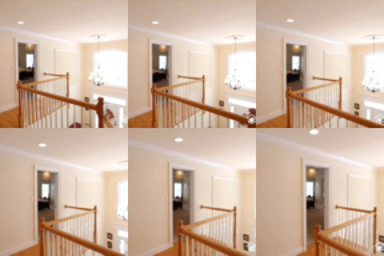}};
                    \end{tikzpicture}
                }
            \end{minipage}
            \begin{minipage}[c]{0.39\textwidth}
                \resizebox{\textwidth}{!}{%
                    \begin{tikzpicture}
                      \node[inner sep=0] (image1) {\includegraphics[height=2cm]{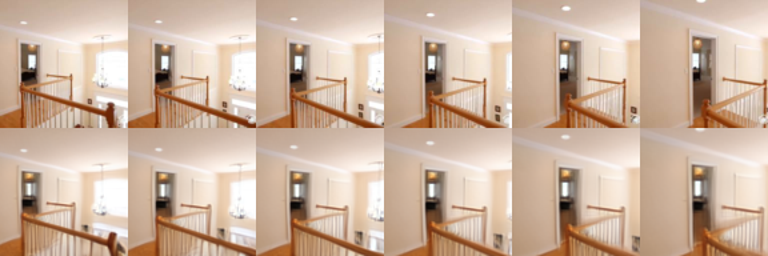}};
                    \end{tikzpicture}
                }
            \end{minipage}
            \begin{minipage}[c]{0.38\textwidth}
                \resizebox{\textwidth}{!}{%
                    \begin{tikzpicture}
                      \node[inner sep=0] (image1) {\includegraphics[height=2cm, clip, trim=61.3cm 0 0 0]{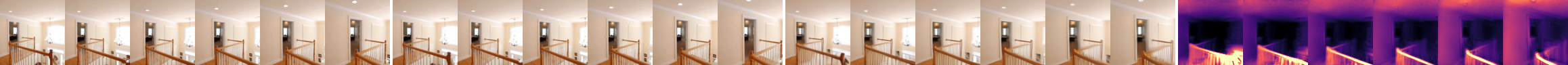}};
                    \end{tikzpicture}
                }
            \end{minipage}
    }
    \resizebox{\textwidth}{!}{%
            \begin{minipage}[c]{0.2\textwidth}
                \resizebox{\textwidth}{!}{%
                    \begin{tikzpicture}
                      \node[inner sep=0] (image1) {\includegraphics[height=2cm]{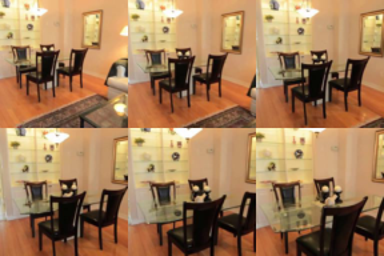}};
                    \end{tikzpicture}
                }
            \end{minipage}
            \begin{minipage}[c]{0.39\textwidth}
                \resizebox{\textwidth}{!}{%
                    \begin{tikzpicture}
                      \node[inner sep=0] (image1) {\includegraphics[height=2cm]{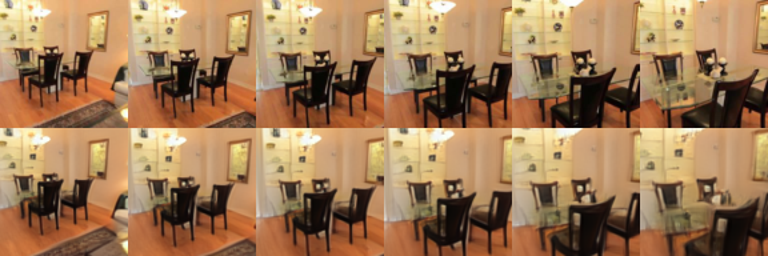}};
                    \end{tikzpicture}
                }
            \end{minipage}
            \begin{minipage}[c]{0.38\textwidth}
                \resizebox{\textwidth}{!}{%
                    \begin{tikzpicture}
                      \node[inner sep=0] (image1) {\includegraphics[height=2cm, clip, trim=61.3cm 0 0 0]{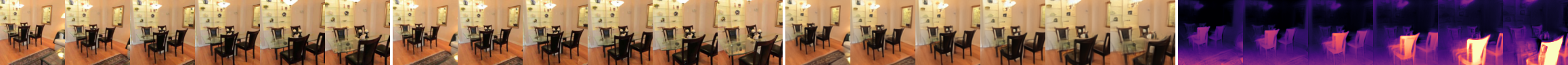}};
                    \end{tikzpicture}
                }
            \end{minipage}
    }
    \caption{RealEstate10k}
    \label{subfig:6-im-recon-ra10k-app}
    \end{subfigure}
    \caption{Additional qualitative results for 3D reconstruction from six images from our model on MVImgNet (a) and RealEstate10k (b) datasets. First column shows input (conditioning) images, second column shows the ground truth images (above) and images rendered from the sampled 3D scene (below), while the third column shows its corresponds depths.}
    \label{fig:6-im-recon-app}
\end{figure}

\end{document}